\documentclass{article}

\usepackage{PRIMEarxiv}

\usepackage[utf8]{inputenc} 
\usepackage[T1]{fontenc}    
\usepackage{hyperref}       
\usepackage{graphicx,booktabs,multirow,makecell}
\usepackage{subfigure} 
\usepackage{algorithmic}
\usepackage{array}
\usepackage{algorithm}
\usepackage{utfsym}
\usepackage{xcolor}
\usepackage{multirow}
\usepackage{graphicx,verbatim}
\usepackage{amsmath}   
\usepackage{amssymb}    
\usepackage{bm}         
\usepackage{mathtools}
\usepackage{booktabs}   
\usepackage{graphicx}
\usepackage{hyperref}
\usepackage{algorithm}
\usepackage{algorithmic}
\usepackage{multirow}
\usepackage{tabularx}
\usepackage{colortbl}
\usepackage{array}
\usepackage{natbib}
\usepackage{ulem}     
\graphicspath{{media/}}     
\usepackage{authblk}
\title{Information bottleneck-guided heterogeneous graph learning for interpretable neurodevelopmental disorder diagnosis
\thanks{\textit{\underline{Citation}}: 
\textbf{Y Li, L Chen, W Dong, S Gong, Z Kang, B Wei, W Zeng, H Yan, L Bian, Z Zhang, WT Siok, N Wang. Information bottleneck-guided heterogeneous graph learning for interpretable neurodevelopmental disorder diagnosis [J]. Neurocomputing, 2026, DOI:10.1016/j.neucom.2026.133721.}} 
}

\author[1,2,$\dagger$]{Yueyang Li}
\author[2,$\dagger$]{Lei Chen}
\author[2]{Wenhao Dong}
\author[2]{Shengyu Gong}
\author[2]{Zijian Kang}
\author[2]{Boyang Wei}
\author[2,*]{Weiming Zeng}
\author[3]{Hongjie Yan}
\author[1,4]{Lingbin Bian}
\author[5]{Zhiguo Zhang}
\author[1,*]{Wai Ting Siok}
\author[1,*]{Nizhuan Wang}

\affil[1]{Department of Language Science and Technology, The Hong Kong Polytechnic University, Hung Hom, Kowloon, Hong Kong SAR, China}
\affil[2]{Laboratory of Digital Image and Intelligent Computation, Shanghai Maritime University, Shanghai 201306, China}
\affil[3]{Department of Neurology, Affiliated Lianyungang Hospital of Xuzhou Medical University, Lianyungang 222002, China}
\affil[4]{The State Key Laboratory of Brain and Cognitive Sciences, The University of Hong Kong, Pokfulam, Hong Kong SAR, China}
\affil[5]{Institute of Computing and Intelligence, Harbin Institute of Technology Shenzhen, Shenzhen 518000, China}
\affil[$\dagger$]{Co-first authors}
\affil[*]{Correspondence: wmzeng@shmtu.edu.cn; wai-ting.siok@polyu.edu.hk; wangnizhuan1120@gmail.com}

\begin{document}
\maketitle
\begin{abstract}
Developing interpretable models for neurodevelopmental disorders (NDDs) diagnosis presents significant challenges in effectively encoding, decoding, and integrating multimodal neuroimaging data. While many existing machine learning approaches have shown promise in brain network analysis, they typically suffer from limited interpretability, particularly in extracting meaningful biomarkers from functional magnetic resonance imaging (fMRI) data and establishing clear relationships between imaging features and demographic characteristics. Besides, current graph neural network methodologies face limitations in capturing both local and global functional connectivity patterns while simultaneously achieving theoretically principled multimodal data fusion. To address these challenges, we propose the Interpretable Information Bottleneck Heterogeneous Graph Neural Network (I²B-HGNN), a unified framework that applies information bottleneck principles to guide both brain connectivity modeling and cross-modal feature integration. This framework comprises two complementary components. The first is the Information Bottleneck Graph Transformer (IBGraphFormer), which combines transformer-based global attention mechanisms with graph neural networks through information bottleneck-guided pooling to identify sufficient biomarkers. The second is the Information Bottleneck Heterogeneous Graph Attention Network (IB-HGAN), which employs meta-path-based heterogeneous graph learning with structural consistency constraints to achieve interpretable fusion of neuroimaging and demographic data. The experimental results demonstrate that I²B-HGNN achieves superior performance in diagnosing NDDs, exhibiting both high classification accuracy and the ability to provide interpretable biomarker identification while effectively analyzing non-imaging data. The official implementation code is published at https://github.com/RyanLi-X/I2B-HGNN.
\end{abstract}
\keywords{Information Bottleneck, Interpretability, Multimodal, Brain Network, Neurodevelopmental disorder, Heterogeneous Graph, Functional magnetic resonance imaging (fMRI).}

\section{Introduction}
Neurodevelopmental disorders (NDDs), such as attention deficit hyperactivity disorder (ADHD) and autism spectrum disorder (ASD), significantly impact cognitive and social development, posing major challenges for affected individuals \cite{thapar2017neurodevelopmental,dogra2025development}. Traditional behavioral assessments rely on subjective clinical observations and standardized questionnaires, leading to inherent variability in diagnosis \cite{li2021braingnn,han2024multi,gao2025integrating}. These inconsistencies frequently delay interventions, particularly for ASD where early identification is crucial \cite{dong2024starformer}, and are compounded by the heterogeneity of NDDs wherein individuals with the same diagnosis often demonstrate divergent neurobiological profiles \cite{li2025mhnet}, highlighting the urgent need for objective, data-driven diagnostic tools. Functional magnetic resonance imaging (fMRI) has emerged as a powerful neuroimaging modality for objective assessment of brain function in psychiatric disorders \cite{mubonanyikuzo2025detection,li2025mhnet,mukhtar2025design,mukhtar2025novel}, offering non-invasive measurement of neural activity through blood-oxygen-level-dependent (BOLD) signal detection \cite{zhang2024stanet,feng2024neural}. Unlike structural imaging techniques, fMRI captures dynamic brain connectivity patterns that directly reflect underlying neurobiological processes \cite{cai2024mm,raja2025design,raja2025hybrid}. This technique's capacity to quantify functional connectivity (FC) between brain regions enables the development of objective, data-driven diagnostic tools that can complement traditional clinical assessments. However, developing interpretable diagnostic models remains challenging, as it requires careful balance between clinically meaningful biomarker interpretation and effective integration of multimodal imaging and behavioral data.

Although graph neural networks (GNNs) have shown promise in analyzing functional brain connectomes \cite{parisot2018disease}, many existing approaches face fundamental limitations. While brain connectivity-based GNNs effectively extract biomarkers, they often suffer from limited generalizability due to single-site training \cite{xu2024contrastive}, whereas population graph approaches, despite improving accuracy through phenotypic similarity modeling, may compromise the discovery of biologically meaningful connectivity patterns crucial for understanding psychiatric disorders \cite{cai2024mm,insel2015brain}. Conventional homogeneous graph models restrict the use of non-imaging data by only mapping demographic information to edge weights, limiting their ability to fully utilize the rich multimodal information available in clinical settings \cite{wang2019heterogeneous}. While GNNs can effectively model local connectivity patterns through neighborhood aggregation, their limited receptive field restricts the ability to capture long-range dependencies that span multiple brain networks for identifying distributed biomarkers \cite{li2021braingnn}. Conversely, transformer-based models excel at capturing unconstrained dependencies between any pair of brain regions through self-attention mechanisms. However, unlike GNNs, they lack the inherent capacity to incorporate neuroanatomical priors or preserve brain network topology \cite{dong2024starformer}. Indeed, most existing approaches lack principled frameworks for balancing model complexity with interpretability, frequently resulting in models that either compromise performance or clinical utility \cite{zhang2022classification,shao2023heterogeneous}.

Current interpretability approaches present significant limitations in neuroimaging, often failing to reveal meaningful cross-modal interactions due to detachment from actual decision processes \cite{hemker2025healnet}, typically providing feature importance scores without considering the complex interdependencies between imaging and non-imaging variables crucial for understanding psychiatric disorders. Moreover, existing diagnostic models struggle to identify minimal sufficient representations that are most informative for classification while eliminating redundant information \cite{shamir2010learning}. The information bottleneck (IB) principle \cite{tishby2000information} addresses these challenges by enabling optimal compression of connectivity patterns while preserving diagnostically relevant information, establishing a mathematically rigorous framework that simultaneously enhances model generalizability and interpretability \cite{yu2021recognizing}, particularly suitable for medical applications requiring both accuracy and decision transparency. Importantly, accumulating neurobiological evidence demonstrates that brain connectivity patterns fundamentally contribute to both the clinical diagnosis of psychiatric disorders and the characterization of their neural substrates \cite{insel2015brain,raza2025stochastic}.

To systematically address these challenges, we present the Interpretable Information Bottleneck Heterogeneous Graph Neural Network (I²B-HGNN), which operates at two complementary levels: (1) individual-level brain network analysis through the Information Bottleneck GraphFormer (IBGraphFormer), and (2) population-level heterogeneous graph learning via the Information Bottleneck Heterogeneous Graph Attention Network (IB-HGAN). This dual-level approach enables comprehensive understanding of both subject-specific neural patterns and population-wide diagnostic characteristics. Overall, we present three main contributions as follows:
\begin{enumerate}
\renewcommand{\labelenumi}{\arabic{enumi})}
	\item \textbf{Integrated IB Framework}: We propose a unified architecture that applies IB principles to both brain connectivity modeling and multimodal fusion. This framework identifies minimal sufficient biomarkers while preserving essential cross-modal interactions between imaging and non-imaging features, effectively addressing the accuracy-interpretability trade-off through theoretical guidance.
	\item \textbf{Interpretable Biomarker Identification}: The IBGraphFormer synergistically combines transformer-based global attention with GNNs through IB-guided pooling. This design enables interpretable biomarker extraction by learning minimal sufficient statistics that preserve both local and global FC patterns essential for diagnostic decisions.
	\item \textbf{Theoretically Principled Multimodal Integration}: The IB-HGAN employs an information-theoretic approach to heterogeneous graph learning, utilizing meta-path-based population graphs and graph isomorphism testing to ensure structural consistency. This component enables explicit attribution of both imaging and non-imaging features while maintaining neurobiologically valid cross-modal interactions for enhanced diagnostic interpretability.
\end{enumerate}

The remainder of the paper is organized as follows. Section \ref{Related Work} reviews research work on GNNs and interpretable learning approaches. Section \ref{Method} present our proposed IBGraphFormer and IB-HGAN modelling framework. Section \ref{Experiments} presents the experimental results and analyses on publicly available datasets and explores the contributing brain regions. The detailed experimental setup including datasets and competing methods is discussed and analysed in Section \ref{Result}. Section \ref{Discussion} discusses feature visualization, parameter analysis, interpretability analysis, limitation and future work. Section \ref{conclusion} draws the conclusion.

\section{Related Work}\label{Related Work}
\subsection{GNNs for Brain Connectivity Analysis}
NDDs diagnosis has traditionally relied on manually extracted features from fMRI data combined with conventional machine learning classifiers \cite{li2021braingnn,parisot2018disease}. While early methods focused on statistical measures of functional connectivity and graph-theoretic properties, they often failed to capture complex topological patterns in brain networks \cite{yan2024review,zhao2022dynamic}. The introduction of GNNs has revolutionized brain network analysis by enabling end-to-end learning of graph representations \cite{kipf2017semi,chen2022adversarial}. Population-based GCN approaches model inter-subject relationships, where subjects serve as nodes and phenotypic similarities define edges \cite{parisot2018disease,kazi2019inceptiongcn,huang2022disease}. Although effective for leveraging population-level information, these methods face scalability issues and provide limited insights into individual brain connectivity patterns \cite{bessadok2022graph}. Brain connectome-based GNNs directly model individual brain networks, treating regions of interest (ROIs) as nodes and FC strengths as edge weights \cite{li2021braingnn}. Recent developments have explored residual connections for ASD prediction \cite{wang2024residual} and dual-view connectivity approaches that capture complementary brain network representations \cite{guan2024dynamic}. However, existing brain GNN approaches typically assume homogeneous graph structures and fail to capture long-range dependencies across multiple brain systems \cite{dong2024starformer}. More critically, existing methods lack principled frameworks for identifying minimal sufficient biomarkers while maintaining interpretability, limiting their clinical translation potential.

\subsection{IB and Interpretable Learning}
The IB principle provides a theoretical framework for learning compressed representations that preserve task-relevant information while discarding irrelevant details \cite{tishby2000information}. The IB objective seeks to find representations $T$ that minimize $I(X;T)$ while maximizing $I(Y;T)$, where $I(\cdot;\cdot)$ denotes mutual information \cite{shamir2010learning}. Recent surveys highlight the growing importance of IB theory across various domains \cite{hu2024survey}. In graph learning, subgraph IB methods have been introduced for identifying predictive substructures \cite{yu2021recognizing}. The application of IB principles to medical imaging and brain network analysis has gained significant momentum, with recent work including BrainIB for interpretable psychiatric diagnosis while identifying meaningful biomarkers \cite{zheng2024brainib}, and dynamic graph attention IB for temporal brain network analysis \cite{dong2024brain}. These methods successfully balance compression and prediction performance but typically focus on single-modal neuroimaging data. Current IB-based approaches also lack comprehensive frameworks for handling heterogeneous information sources and multimodal data fusion, which are crucial for capturing the full complexity of psychiatric disorders \cite{insel2015brain}.

\subsection{Multimodal and Heterogeneous Graph Learning}
Recent advances in heterogeneous graph learning offer promising solutions for multimodal integration by explicitly modeling different types of nodes and edges to capture diverse data relationships. Heterogeneous graph attention networks have been developed for knowledge graphs \cite{wang2019heterogeneous}, though adapting these methods to weighted brain networks presents significant challenges due to the continuous nature of neuroimaging features and the need to preserve neuroanatomical constraints. Early neuroimaging applications explored heterogeneous approaches for integrating structural and FC data, primarily focusing on simple node type distinctions rather than comprehensive multimodal fusion \cite{huang2020attention}. Recent developments have made substantial progress in addressing these limitations. FC-HGNN introduces brain hemisphere-based heterogeneous modeling for mental disorder identification, demonstrating improved performance through explicit modeling of inter- and intra-hemispheric connections \cite{gu2025fc}. IFC-GNN incorporates FC interactions with multimodal GNNs, enabling more sophisticated cross-modal feature learning \cite{wang2024ifc}. Chen et al. exploring multi-connectivity patterns through heterogeneous graphs that capture both global and local brain network properties \cite{chen2024exploring}. Despite these advances, existing approaches predominantly rely on heuristic fusion strategies without theoretical guidance for optimal information integration, and they often lack interpretable mechanisms for understanding cross-modal feature interactions.

\section{Method}\label{Method}
\subsection{Problem Formulation}
\begin{figure}[t]
	\includegraphics[width=\textwidth]{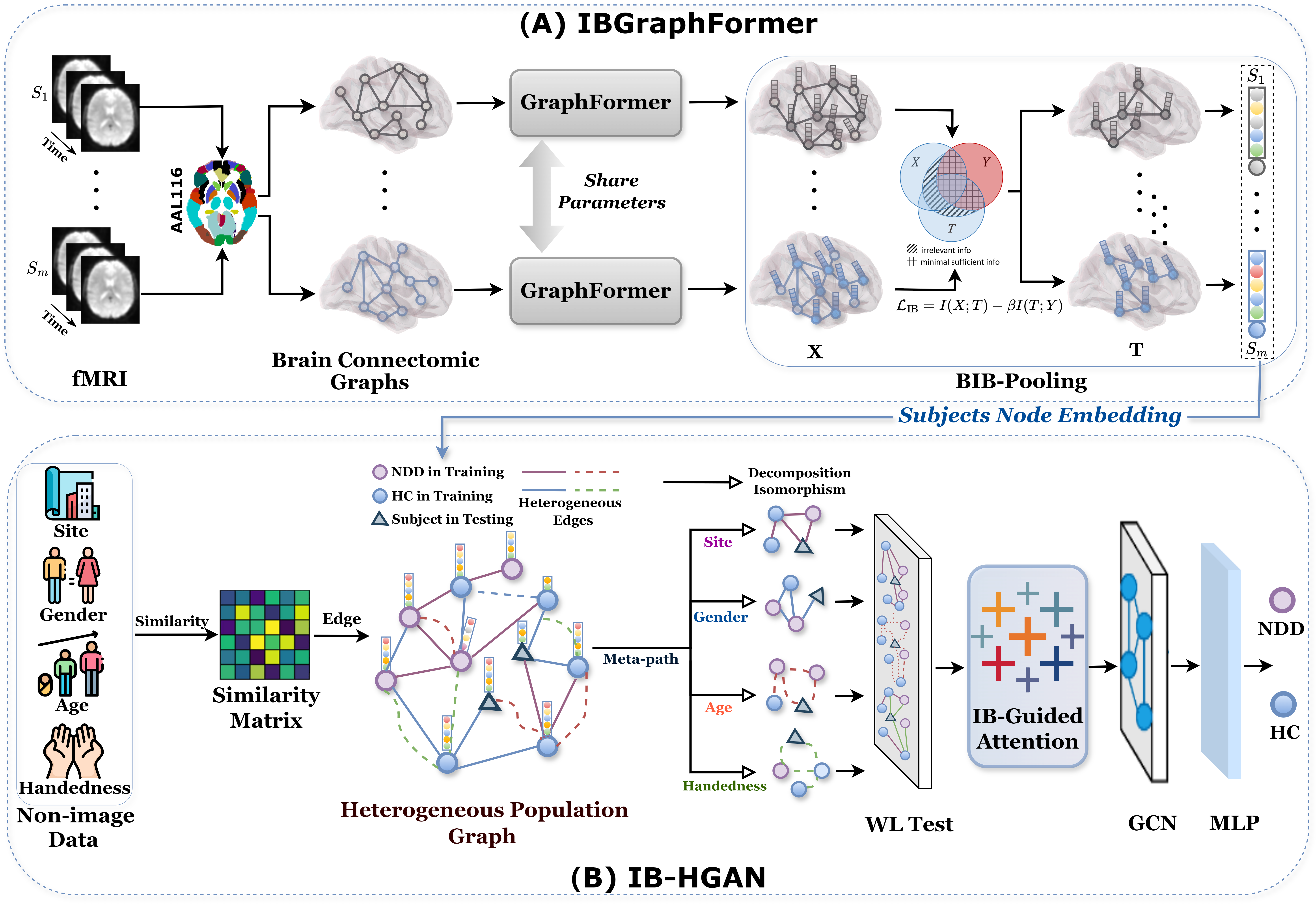}
	\caption{Illustration of our I²B-HGNN for NDDs diagnosis. (A) IBGraphFormer performs individual-level brain network analysis on Brain Connectomic Graphs as individual graph embeddings through distribution-aware global attention and BIB-pooling for biomarker extraction. (B) IB-HGAN conducts population-level heterogeneous graph learning on Heterogeneous Population Graphs as demographic-based heterogeneous graphs for NDDs classification using meta-path attention with structural consistency constraints for interpretable multimodal fusion.} \label{model}
\end{figure}
Given preprocessed fMRI data $\mathbf{X}_i \in \mathbb{R}^{|\mathcal{V}| \times T}$ for the $i$-th subject, where $T$ denotes the number of time points, and a set of brain connectome graphs $\mathcal{G} = \{G_1, G_2, \ldots, G_N\}$ derived from these time series, where each graph $G_i = (\mathcal{V}, \mathcal{E}, \mathbf{A}_i)$ represents the $i$-th subject with $\mathcal{V}$ denoting brain regions as nodes, $\mathcal{E}$ representing functional connections as edges, and $\mathbf{A}_i \in \mathbb{R}^{|\mathcal{V}| \times |\mathcal{V}|}$ being the FC matrix, we aim to learn a mapping function $f: (\mathcal{G}, \mathbf{D}) \rightarrow \mathcal{Y}$ that accurately predicts NDDs labels $\mathbf{y} \in \{0, 1\}^N$, where $\mathbf{D} = [\mathbf{d}_1, \mathbf{d}_2, \ldots, \mathbf{d}_N]^T$ represents demographic information including age, gender, site, and handedness. The FC matrix is computed as $\mathbf{A}_i[j,k] = \text{corr}(\mathbf{X}_i[j,:], \mathbf{X}_i[k,:])$ representing Pearson correlations between ROIs time series and then applying fisher z-transformation for normalization.

\subsection{IBGraphFormer}
\begin{figure}[t]
	\includegraphics[width=\textwidth]{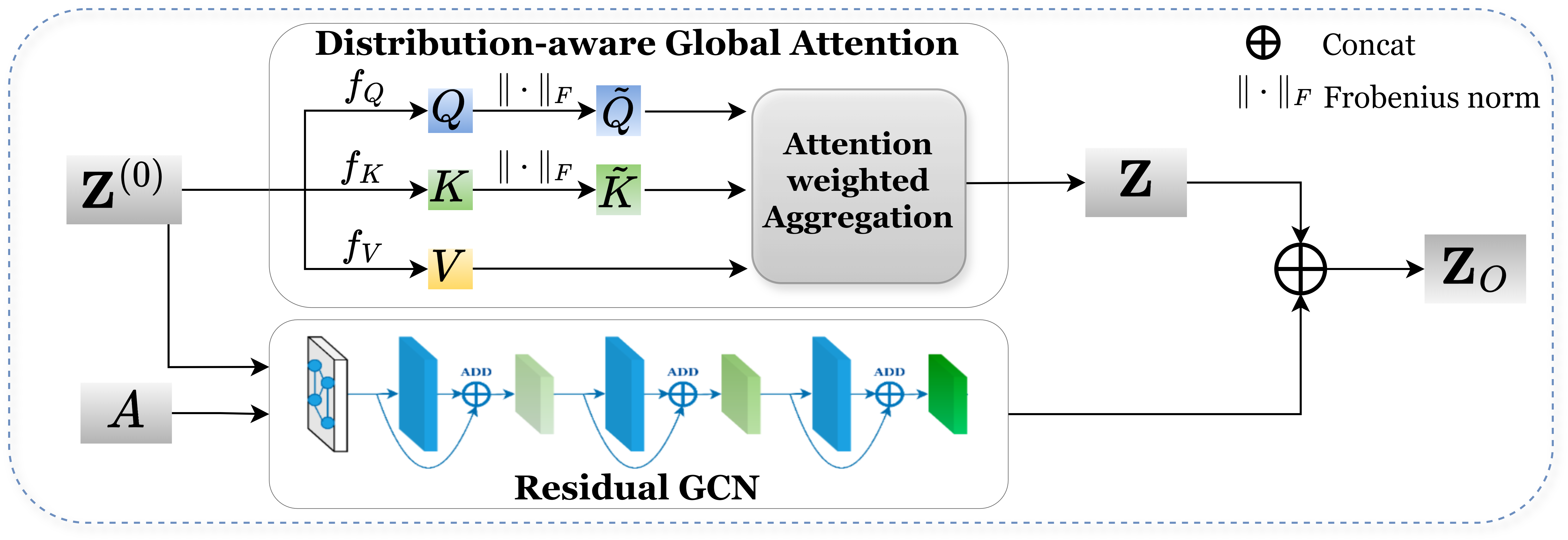}
	\caption{Illustration of Distribution-aware Global Attention GraphFormer.} \label{Former}
\end{figure}
\subsubsection{Distribution-aware Global Attention GraphFormer}
To capture intrinsic FC patterns, we first employ a neural mapping layer $f_I$ that projects input features into latent embeddings $\mathbf{Z}^{(0)} = f_I(\mathbf{X})$, where $\mathbf{Z}^{(0)} \in \mathbb{R}^{|\mathcal{V}| \times d}$ serves as the initial node features with $d$ being the embedding dimension. The Distribution-aware Global Attention GraphFormer captures distributed connectivity patterns across brain networks, facilitating the modeling of long-range dependencies that span multiple functional networks. We employ a single-layer global attention design for brain network analysis, as one-layer propagation over densely connected attention graphs enables adaptive information propagation between arbitrary node pairs \cite{wu2019simplifying}. The distribution-aware attention design explicitly models the statistical properties of attention weights, addressing the issue that attention weights become overly concentrated on a few dominant regions, thereby limiting the model's ability to capture distributed patterns characteristic of brain networks. The global attention module quantifies cross-ROI influences as:
\begin{equation}
    \{\mathbf{Q}, \mathbf{K}, \mathbf{V}\} = \{f_Q, f_K, f_V\}(\mathbf{Z}^{(0)}), \quad
    \{\tilde{\mathbf{Q}}, \tilde{\mathbf{K}}\} = \{\mathbf{Q}/\|\mathbf{Q}\|_F, \mathbf{K}/\|\mathbf{K}\|_F\}
\end{equation}
where $f_Q$, $f_K$, $f_V$ denote learnable feature transformation functions, and $\|\cdot\|_F$ denotes the Frobenius norm for normalizing attention distributions. This normalization strategy maintains the attention distribution within a bounded range, preventing the attention collapse phenomenon.

The attention-weighted feature aggregation process is formulated as:
\begin{equation}
\mathbf{Z}=\lambda\mathbf{N}^{-1}\left[\mathbf{V}+\frac{1}{|\mathcal{V}|}\tilde{\mathbf{Q}}(\tilde{\mathbf{K}}^{\top}\mathbf{V})\right]+(1-\lambda)\mathbf{Z}^{(0)}
\end{equation}
where $\lambda$ is a learnable parameter that balances between global attention and original node features. The normalization matrix $\mathbf{N} = \text{diag}\left(1 + \frac{1}{|\mathcal{V}|}\tilde{\mathbf{Q}}(\tilde{\mathbf{K}}^{\top}\mathbf{e})\right)$ serves as a degree normalization term that prevents over-smoothing by accounting for the effective attention degree of each node, where $\mathbf{e} \in \mathbb{R}^{|\mathcal{V}|}$ is the all-ones vector and $\text{diag}(\cdot)$ constructs a $|\mathcal{V}| \times |\mathcal{V}|$ diagonal matrix. The single-layer design reduces computational complexity from $\mathcal{O}(L \cdot |\mathcal{V}|^2 \cdot d)$ to $\mathcal{O}(|\mathcal{V}|^2 \cdot d)$ compared to $L$-layer attention architectures, while maintaining representational capacity through the global attention mechanism's ability to model arbitrary pairwise dependencies in a single forward pass. The average shortest path length between regions is typically small, making single-layer global attention sufficient for capturing relevant dependencies.

In contrast to the unrestricted global attention mechanism, the residual GNN preserves the original graph structural information by operating within the brain connectivity topology. This component captures local neighborhood structures and maintains the connectivity patterns inherent in the functional brain network through message passing along existing connections:
\begin{equation}
\text{Residual GNN}(\mathbf{Z}^{(0)}, \mathbf{A}) = \sigma\left(\tilde{\mathbf{D}}^{-\frac{1}{2}}\tilde{\mathbf{A}}\tilde{\mathbf{D}}^{-\frac{1}{2}}\mathbf{Z}^{(0)}\mathbf{W}_{gnn}\right)
\end{equation}
where $\tilde{\mathbf{A}} = \mathbf{A} + \mathbf{I}$ is the adjacency matrix with self-loops, $\tilde{\mathbf{D}}_{ii} = \sum_j \tilde{\mathbf{A}}_{ij}$ represents the degree matrix, $\mathbf{W}_{gnn} \in \mathbb{R}^{d \times d}$ is the learnable weight matrix, and $\sigma(\cdot)$ denotes the activation function. This approach ensures that information aggregation follows the connectivity patterns defined by the functional connectivity matrix, complementing the global view provided by the attention mechanism.

The final integrated representation incorporates both global context and local feature preservation through a residual connection that combines global distributed patterns and local structural information:
\begin{equation}
\mathbf{Z}_O = (1-\gamma)\mathbf{Z} + \gamma\text{Residual GNN}(\mathbf{Z}^{(0)}, \mathbf{A})
\end{equation}
where $\gamma$ is a learnable parameter. This integration enables the model to leverage both long-range dependencies captured by global attention and local connectivity patterns preserved by the residual GNN.

\subsubsection{BIB-Pooling}
To identify diagnostically relevant biomarkers from the integrated node representations $\mathbf{Z}_O \in \mathbb{R}^{|\mathcal{V}| \times d}$ obtained from IBGraphFormer, we develop the Biomarker-oriented Information Bottleneck Pooling (BIB-Pooling) layer that performs graph-level compression to extract subject-specific biomarker representations. The IB framework provides a effective approach for extracting minimal sufficient biomarkers that preserve essential brain connectivity information while discarding redundant connectivity patterns. Formally, for input node representations $\mathbf{Z}_O$ and target labels $Y$, the IB principle seeks to find optimal compressed biomarker representations $\mathbf{T} \in \mathbb{R}^{d_T}$ by optimizing the information-theoretic objective:
\begin{equation}
\mathcal{L} = I(\mathbf{Z}_O;\mathbf{T}) - \beta I(\mathbf{T};Y)
\end{equation}
where $I(\cdot;\cdot)$ denotes mutual information, $\beta$ controls the trade-off between compression and predictive accuracy, and $d_T$ represents the dimensionality of the compressed biomarker space.

We implement this principle through a variational approximation by parameterizing the encoding distribution $q_{\phi}(\mathbf{T}|\mathbf{Z}_O)$ as a multivariate Gaussian distribution:
\begin{equation}
q_{\phi}(\mathbf{T}|\mathbf{Z}_O) = \mathcal{N}(\boldsymbol{\mu}_{\phi}(\mathbf{Z}_O), \text{diag}(\boldsymbol{\sigma}^2_{\phi}(\mathbf{Z}_O)))
\end{equation}
where $\boldsymbol{\mu}_{\phi}: \mathbb{R}^{|\mathcal{V}| \times d} \rightarrow \mathbb{R}^{d_T}$ and $\boldsymbol{\sigma}_{\phi}: \mathbb{R}^{|\mathcal{V}| \times d} \rightarrow \mathbb{R}^{d_T}$ are multi-layer perceptrons with global average pooling that transform graph-level node features into mean and variance parameters, respectively. The biomarker representation is sampled using the reparameterization trick:
\begin{equation}
\mathbf{T} = \boldsymbol{\mu}_{\phi}(\mathbf{Z}_O) + \boldsymbol{\sigma}_{\phi}(\mathbf{Z}_O) \odot \boldsymbol{\epsilon}, \quad \boldsymbol{\epsilon} \sim \mathcal{N}(0, \mathbf{I}_{d_T})
\end{equation}
where $\odot$ denotes element-wise multiplication and $\mathbf{I}_{d_T}$ is the $d_T$-dimensional identity matrix.

The practical BIB-Pooling loss function implements the theoretical objective through a variational lower bound:
\begin{equation}
\mathcal{L}_{BIB} = \mathbb{E}_{q_{\phi}(\mathbf{T}|\mathbf{Z}_O)}[-\log p_{\theta}(Y|\mathbf{T})] + \beta \cdot \text{KL}(q_{\phi}(\mathbf{T}|\mathbf{Z}_O)||p(\mathbf{T}))
\end{equation}
where the first term maximizes biomarker predictive power through a parameterized classifier $p_{\theta}(Y|\mathbf{T})$, and the KL divergence term enforces complexity regularization against a standard Gaussian prior $p(\mathbf{T}) = \mathcal{N}(0, \mathbf{I}_{d_T})$, with $\theta$ and $\phi$ denoting learnable parameters. This theoretically-grounded approach enables automatic identification of sparse yet diagnostically meaningful biomarkers by balancing predictive relevance with model complexity through principled information compression.

\subsection{IB-HGAN}
\subsubsection{Heterogeneous Population Graph Construction}
Building upon the biomarker representations $\{\mathbf{T}_i\}_{i=1}^N$ extracted by IBGraphFormer, we construct a heterogeneous population graph $\mathcal{G}_H = (\mathcal{V}_H, \mathcal{E}_H, \mathcal{R})$ to model complex inter-subject relationships and achieve interpretable multimodal integration. During training, the population graph is constructed using only the available training subjects, where each subject node $v_i \in \mathcal{V}_H$ is associated with its biomarker representation $\mathbf{T}_i$ and demographic features $\mathbf{d}_i = [d_i^{\text{site}}, d_i^{\text{sex}}, d_i^{\text{age}}, d_i^{\text{hand}}]$. We define four relationship types $\mathcal{R} = \{r_{\text{site}}, r_{\text{sex}}, r_{\text{age}}, r_{\text{hand}}\}$ and construct corresponding meta-path subgraphs $\{\mathcal{G}_k\}_{k=1}^4$ with adjacency matrices $\{\mathbf{A}_k\}_{k=1}^4$, where each meta-path captures specific demographic relationships while preserving the heterogeneity of population-level interactions.

The demographic compatibility function is defined as:
\begin{equation}
C_{k}(d_i^k, d_j^k) = \begin{cases}
1 & \text{if } d_i^k = d_j^k \text{ (site, sex, handedness)} \\
\exp\left(-\frac{|d_i^k - d_j^k|^2}{2\sigma_k^2}\right) & \text{if } |d_i^k - d_j^k| \leq \tau_k \text{ (age)}
\end{cases}
\end{equation}
where $\sigma_k$ controls the Gaussian kernel bandwidth for age similarity and $\tau_k$ defines the connection threshold. For each meta-path $k$, the edge weights incorporate biomarker similarity with demographic compatibility:
\begin{equation}
\mathbf{A}_k[i,j] = \text{Sim}(\mathbf{T}_i, \mathbf{T}_j) \odot C_{k}(d_i^k, d_j^k)
\end{equation}
where $\odot$ denotes element-wise multiplication. The biomarker similarity is computed as:
\begin{equation}
\text{Sim}(\mathbf{T}_i, \mathbf{T}_j) = \exp\left(-\frac{[\rho(\mathbf{T}_i, \mathbf{T}_j)]^2}{2\sigma^2}\right)
\end{equation}
where $\rho(\cdot)$ represents the distance function and $\sigma$ denotes the kernel bandwidth.

\subsubsection{Meta-Path Structural Consistency Learning}
Different demographic attributes may result in topologically similar population structures that should receive equivalent attention weights to maintain model interpretability and stability. We employ the Weisfeiler-Lehman (WL) graph isomorphism test to systematically identify meta-paths with equivalent structural properties through iterative neighborhood aggregation, ensuring consistent treatment of structurally equivalent meta-paths in heterogeneous graph learning. The WL algorithm iteratively refines node labels based on neighborhood structures:
\begin{equation}
\ell_i^{(t+1)} = \text{HASH}(\ell_i^{(t)}, \{\ell_j^{(t)} : j \in \mathcal{N}(i)\})
\end{equation}
where $\ell_i^{(t)}$ denotes the label of node $i$ at iteration $t$, $\mathcal{N}(i)$ represents the neighborhood of node $i$, and $\text{HASH}(\cdot)$ is a deterministic hash function that assigns unique labels to distinct neighborhood patterns. The initial labels $\ell_i^{(0)}$ are set uniformly for all nodes since we focus on structural rather than attribute-based equivalence.

This iterative process captures complex structural patterns beyond simple degree-based comparisons, enabling the detection of higher-order topological equivalences between meta-paths. The structural equivalence between meta-paths $\mathcal{G}_i$ and $\mathcal{G}_j$ is quantified through the WL isomorphism test:
\begin{equation}
S_{ij} = \mathbb{I}[\mathcal{I}(\mathcal{G}_i, \mathcal{G}_j)]
\end{equation}
where $\mathbb{I}[\cdot]$ is the indicator function and $\mathcal{I}(\cdot, \cdot)$ denotes the graph isomorphism test that returns true if two graphs are structurally equivalent based on the WL algorithm. Meta-paths with $S_{ij} = 1$ are considered structurally equivalent. This structural equivalence matrix $\mathbf{S} \in \{0,1\}^{4 \times 4}$ becomes the foundation for constraining attention learning in the subsequent attention mechanism, ensuring that topologically identical meta-paths receive consistent treatment.

\subsubsection{IB Guided Heterogeneous Graph Meta-Path Attention}
Conventional attention mechanisms in heterogeneous graphs rely on heuristic similarity measures without theoretical guidance for optimal information selection. By integrating IB principles into meta-path attention learning, our approach provides theoretical guarantees for achieving minimal sufficient statistics while preserving predictive performance. The fundamental insight is that each meta-path should contribute to the final representation proportionally to its diagnostic relevance, while redundant demographic pathways should be suppressed through principled information compression. Algorithm \ref{alg1} shows the procedure of IB guided heterogeneous graph meta-path attention.

For each meta-path $k$, we employ Graph Attention Networks to learn path-specific node representations $\mathbf{Z}_k = \text{GAT}(\mathbf{T}, \mathbf{A}_k)$, where $\mathbf{T} = \{\mathbf{T}_i\}_{i=1}^N$ denotes the biomarker representations extracted from IBGraphFormer. The theoretical foundation for determining meta-path importance lies in quantifying the mutual information between biomarker features and path-specific representations. We employ the Donsker-Varadhan representation to estimate the mutual information:
\begin{equation}
I(\mathbf{T};\mathbf{Z}_k) \approx \sup_{\psi_k} \mathbb{E}_{p(\mathbf{T},\mathbf{Z}_k)}[\psi_k(\mathbf{t},\mathbf{z})] - \log\mathbb{E}_{p(\mathbf{T})p(\mathbf{Z}_k)}[e^{\psi_k(\mathbf{t},\mathbf{z})}]
\end{equation}
where $\psi_k$ is a neural network discriminator that distinguishes joint distributions from marginal distributions, providing a tractable approximation for mutual information estimation in high-dimensional spaces.

The IB-guided attention weight computation incorporates both representational capacity and information compression constraints, where $\mathbf{u} \in \mathbb{R}^{d_{att}}$, $\mathbf{W}_{\text{att}} \in \mathbb{R}^{d_{att} \times d}$, $\mathbf{b}_{\text{att}} \in \mathbb{R}^{d_{att}}$ are learnable parameters with $d_{att}$ being the attention hidden dimension, and $\beta_H$ controls information compression strength. The preliminary attention scores $\hat{\alpha}_k$ are computed through the procedure detailed in Algorithm \ref{alg1}, where the exponential term implements the IB principle by penalizing meta-paths with excessive mutual information, promoting selective attention to diagnostically relevant pathways while suppressing redundant demographic relationships.

To enforce structural consistency across equivalent meta-paths, we utilize the structural equivalence matrix $\mathbf{S}$ derived from the WL test. For meta-paths identified as structurally equivalent ($S_{ij} = 1$), we enforce the constraint $\alpha_i = \alpha_j$ to ensure consistent treatment. This constraint ensures that meta-paths with equivalent topological configurations receive identical attention weights, preventing arbitrary attention variations while maintaining the theoretical foundation established through graph isomorphism testing. The heterogeneous representation synthesis integrates information from all meta-paths through structurally-constrained attention aggregation:
\begin{equation}
\mathbf{Z}_H = \sum_{k=1}^4 \alpha_k\mathbf{Z}_k
\end{equation}
where the attention weights $\boldsymbol{\alpha}$ have been optimized under both information-theoretic principles and structural equivalence constraints, ensuring minimal sufficient statistics while maintaining consistency across topologically equivalent meta-paths.

\begin{algorithm}[t]
\caption{IB Guided Heterogeneous Graph Meta-Path Attention}
\label{alg1}
\begin{algorithmic}[1]
\REQUIRE $\{\mathbf{T}_i\}_{i=1}^N$, $\{\mathbf{A}_k\}_{k=1}^4$, $\mathbf{S}$
\ENSURE $\mathbf{Z}_H$
\FOR{each meta-path $k = 1$ to $4$}
\STATE $\mathbf{Z}_k = \text{GAT}(\mathbf{T}, \mathbf{A}_k)$
\STATE $\mathbf{z}_k = \frac{1}{N}\sum_{i=1}^N \mathbf{Z}_k[i,:]$
\STATE Estimate $I(\mathbf{T};\mathbf{Z}_k)$ using discriminator $\psi_k$
\STATE $\hat{\alpha}_k = \mathbf{u}^{\top}\tanh(\mathbf{W}_{\text{att}}\mathbf{z}_k + \mathbf{b}_{\text{att}}) \cdot \exp(-\beta_H \cdot I(\mathbf{T};\mathbf{Z}_k))$
\ENDFOR
\STATE Apply structural consistency constraints: $\alpha_i = \alpha_j$ if $S_{ij} = 1$
\STATE Normalize attention weights: $\alpha_k = \frac{\exp(\hat{\alpha}_k)}{\sum_{l=1}^4 \exp(\hat{\alpha}_l)}$
\STATE Compute heterogeneous representation using Eq. (15)
\RETURN $\mathbf{Z}_H$
\end{algorithmic}
\end{algorithm}

\subsubsection{Adaptive Multimodal Integration}
We formulate heterogeneous graph learning as a constrained variational information bottleneck problem. The encoding distribution is parameterized as a multivariate Gaussian:
\begin{equation}
q_{\phi}(\mathbf{Z}_H|\mathbf{T},\mathbf{D}) = \mathcal{N}(\boldsymbol{\mu}_H(\mathbf{T},\mathbf{D}), \text{diag}(\boldsymbol{\sigma}_H^2(\mathbf{T},\mathbf{D})))
\end{equation}
where $\boldsymbol{\mu}_H: \mathbb{R}^{N \times d} \times \mathbb{R}^{N \times 4} \rightarrow \mathbb{R}^{d_H}$ and $\boldsymbol{\sigma}_H: \mathbb{R}^{N \times d} \times \mathbb{R}^{N \times 4} \rightarrow \mathbb{R}^{d_H}$ are neural networks that transform the concatenated biomarker and demographic features into distribution parameters with $d_H$ being the final representation dimension.

The heterogeneous information bottleneck objective balances predictive performance with representation compression:
\begin{equation}
\mathcal{L}_{\text{HIB}} = \mathbb{E}_{q_{\phi}(\mathbf{Z}_H|\mathbf{T},\mathbf{D})}[-\log p_{\theta}(Y|\mathbf{Z}_H)] + \beta_H \cdot \text{KL}(q_{\phi}(\mathbf{Z}_H|\mathbf{T},\mathbf{D})||p(\mathbf{Z}_H))
\end{equation}
where $p(\mathbf{Z}_H) = \mathcal{N}(0, \mathbf{I}_{d_H})$ serves as a standard Gaussian prior, and $p_{\theta}(Y|\mathbf{Z}_H)$ represents the parameterized diagnostic classifier with learnable parameters $\theta$.

The adaptive integration mechanism incorporates multiple regularization constraints to ensure both theoretical optimality and practical interpretability. The structural consistency term $\mathcal{L}_{\text{struct}} = \sum_{i,j=1}^4 S_{ij}\|\mathbf{Z}_i - \mathbf{Z}_j\|_2^2$ explicitly enforces representation similarity between structurally equivalent meta-paths using the WL-derived equivalence matrix $\mathbf{S}$. The sparsity regularization $\mathcal{L}_{\text{sparse}} = \|\boldsymbol{\alpha}\|_1$ promotes selective attention patterns by encouraging few meta-paths to dominate the final representation. The mutual information regularization $\mathcal{L}_{\text{MI}} = \sum_{k=1}^4 \alpha_k \cdot I(\mathbf{T};\mathbf{Z}_k)$ provides an additional information-theoretic constraint that balances representation capacity with compression efficiency, weighted by the learned attention scores $\alpha_k$.

The complete heterogeneous graph learning objective combines these components:
\begin{equation}
\mathcal{L}_{\text{HG}} = \mathcal{L}_{\text{HIB}} + \mu \mathcal{L}_{\text{struct}} + \kappa \mathcal{L}_{\text{sparse}} + \eta \mathcal{L}_{\text{MI}}
\end{equation}
The complete end-to-end optimization objective synthesizes both individual-level biomarker extraction and population-level multimodal integration:
\begin{equation}
\mathcal{L}_{\text{total}} = \mathcal{L}_{\text{cls}} + \zeta \mathcal{L}_{\text{BIB}} + \omega \mathcal{L}_{\text{HG}}
\end{equation}
where $\mathcal{L}_{\text{cls}}$ represents the cross-entropy classification loss, $\mathcal{L}_{\text{BIB}}$ is the biomarker information bottleneck loss from IBGraphFormer, and $\zeta$, $\omega$ balance different components.

\section{Experiments Setup}\label{Experiments}
\begin{table}
    \small
 \caption{Demographic statistics of the datasets used in this work.\label{data}}
  \centering
  \begin{tabularx}{\textwidth}{>{\centering\arraybackslash}p{2.2cm}>{\centering\arraybackslash}p{1.7cm}>{\centering\arraybackslash}p{1.7cm}>{\centering\arraybackslash}p{2.1cm}>{\centering\arraybackslash}X}
    \toprule
    \textbf{Dataset}	& \textbf{Subgroup}	&\textbf{Number} & \textbf{Gender(M/F)}  & \textbf{Age(mean ± std.)}\\
\midrule
\multirow[m]{2}{*}{ABIDE-I}
& ASD	& 403 & 349/54 & 17.07 ± 7.95 \\
& NC	& 468 & 378/90 & 16.84 ± 7.23 \\
\midrule
\multirow[m]{2}{*}{ADHD-200}
& ADHD	& 218 & 178/39 & 11.56 ± 5.91 \\
& NC	& 364 & 198/166 & 12.42 ± 8.62 \\
    \bottomrule
  \end{tabularx}
\end{table}
\subsection{Datasets and Preprocessing}
We evaluated I²B-HGNN on two publicly accessible neuroimaging datasets for NDDs diagnosis. 

\textbf{ABIDE-I Dataset:} The Autism Brain Imaging Data Exchange I (ABIDE-I) initiative aggregated rs-fMRI and comprehensive phenotypic data from multiple international research centers, creating one of the largest publicly available ASD neuroimaging repositories \cite{di2014autism}. From the original collection of 1,112 subjects across various sites, we followed established research protocols by selecting 871 high-quality subjects that have been consistently used in comparative studies to ensure reproducible results \cite{parisot2018disease}. This refined cohort comprises 403 individuals diagnosed with ASD and 468 neurotypical controls. 

\textbf{ADHD-200 Dataset:} We utilized the ADHD-200 dataset, which contains rs-fMRI scans collected across 8 international research centers \cite{bellec2017neuro}. Our analysis focused on four high-quality acquisition sites: New York University Medical Center, Peking University, Kennedy Krieger Institute, and the University of Pittsburgh, consistent with previous research protocols \cite{zhao2022dynamic}. Following data quality assessment and exclusion of subjects with incomplete information, we obtained a final sample of 582 participants, comprising 218 individuals with ADHD and 364 healthy controls.

\textbf{Preprocessing:} rs-fMRI data preprocessing followed standard neuroimaging protocols, utilizing the C-PAC \cite{craddock2013towards} pipeline for ABIDE-I and the Athena \cite{bellec2017neuro} pipeline for ADHD-200. In particular, the initial five to ten volumes were discarded from each resting-state fMRI scan to account for nonequilibrium magnetization effects. Head motion correction was achieved through rigid-body transformation, with subjects showing motion exceeding 2~mm or $2^\circ$ being excluded from subsequent analysis. To avoid introducing spurious artifacts, we did not implement scrubbing or censoring procedures. Nuisance covariate regression was applied to remove signals from white matter, cerebrospinal fluid (CSF), and six head-motion parameters. Spatial normalization to Montreal Neurological Institute (MNI) standard space was followed by smoothing with a Gaussian kernel of $6 \times 6 \times 6$~mm$^3$ full-width at half-maximum (FWHM). Finally, BOLD signals were band-pass filtered at 0.01--0.1~Hz for ABIDE-I and 0.009--0.08~Hz for ADHD-200 to eliminate high-frequency noise and low-frequency drift.

ROI parcellation was determined using the Anatomical Automatic Labeling (AAL116) atlas \cite{rolls2020automated} and the Schaefer atlas (400 ROIs) \citep{schaefer2018local}. The Schaefer atlas parcels the brain based on functional connectivity, emphasizing functional coherence, while the AAL atlas is organized by anatomical structure, facilitating comparison with traditional neuroanatomical studies. Using both atlases allows us to test our model's robustness under different spatial parcellation schemes. These two atlases are widely used in neuroimaging, making it easier to compare our results with existing work and to enhance interpretability and generalizability. For the Schaefer atlas, we selected the scale of 400 ROIs across seven intrinsic connectivity networks. The AAL atlas partitions the brain anatomically into 116 ROIs. For non-imaging data, categorical variables were encoded in ordinal format while continuous variables were normalized to floating-point values, then concatenated for multimodal integration. The demographic features included gender, age, handedness, and acquisition site, providing complementary information for our multimodal diagnostic framework.

\subsection{Implementation Details and Performance Evaluation} 
I²B-HGNN was implemented in PyTorch and trained on an NVIDIA RTX 2080Ti GPU using the Adam optimizer \cite{kingma2014adam}. The model training employed an initial learning rate of 0.01, weight decay of $5 \times 10^{-3}$, and a maximum of 300 epochs with early stopping with a patience of 30 epochs based on validation performance. Regularization techniques included dropout with rate 0.5 and edge dropout with rate 0.5 to prevent overfitting. All experiments used a fixed random seed of 42 to ensure reproducibility. The loss function hyperparameters were set as $\mu = \kappa = 0.5$, $\eta = 0.2$, $\zeta = 0.5$, $\omega = 0.5$. The IB regularization parameters were optimized specifically for each dataset. For the ABIDE-I dataset, we set $\beta = 0.8$ for IBGraphFormer and $\beta_H = 0.5$ for IB-HGAN. For the ADHD-200 dataset, we employed $\beta = 1.0$ for IBGraphFormer and $\beta_H = 0.8$ for IB-HGAN. These parameters were selected through grid search on the validation set, prioritizing configurations that maximize validation AUC while maintaining interpretable biomarker sparsity (measured by the number of high-weight ROIs).

We evaluated model performance using 10-fold stratified cross-validation and Leave-One-Site-Out (LOSO) validation on both ABIDE-I (HC vs. ASD, 17 sites) and ADHD-200 (HC vs. ADHD, 4 sites) datasets. For 10-fold CV, in each fold, the dataset was partitioned into training, validation, and test sets with an 8:1:1 ratio, and the heterogeneous population graph was constructed exclusively from the training set. For LOSO, each site was iteratively used as the test set while training on all remaining sites, ensuring zero site overlap between training and testing, with the population graph reconstructed for each iteration using only the training sites. Test subjects were evaluated by first extracting their biomarker representations through the trained IBGraphFormer, then incorporating them into the established population graph structure for inference. The evaluation metrics included accuracy (ACC), area under the receiver operating characteristic curve (AUC), and F1 score, with results reported as the mean and standard deviation across all 10 folds or all sites.

To assess statistical significance of performance differences, we conduct 
pairwise comparisons on cross-validation metrics between our method and baseline approaches. For each comparison, we first test the normality of paired differences. If normality holds, a two-sided paired t-test is used; otherwise, we apply the two-sided Wilcoxon signed-rank test as a nonparametric alternative. This adaptive approach ensures robust conclusions regardless of underlying data distributions. All tests are performed using MATLAB (ttest and signrank, R2023b, Mathworks Inc., USA) with significance level set at $p < 0.05$.

\subsection{Competing Methods}
To comprehensively evaluate the performance of I²B-HGNN, we compared it against state-of-the-art methods across different categories. Additionally, we included traditional machine learning baselines such as Support Vector Machine (SVM) and Multi-Layer Perceptron (MLP) for comprehensive comparison.

\textbf{Brain Connectomic-Graph Models (B.GCN)} operate on individual brain connectivity networks, treating brain regions as nodes and FC as edges. This category includes BrainGNN~\cite{li2021braingnn}, which employs graph pooling for interpretable brain network analysis; ContrastPool~\cite{xu2024contrastive}, which uses contrastive learning for explainable brain network classification; RGTNet~\cite{wang2024residual}, which combines residual connections with graph transformers; DGCN~\cite{zhao2022dynamic}, which models dynamic brain connectivity; and AL-NEGAT~\cite{chen2022adversarial}, which applies adversarial learning for multimodal integration.

\textbf{Population-Graph Models (P.GCN)} construct graphs where subjects serve as nodes and phenotypic similarities define edges, capturing population-level relationships. This category includes InceptionGCN~\cite{kazi2019inceptiongcn}, which applies inception modules to population graphs; LG-GNN~\cite{zhang2022classification}, which uses local-to-global GNNs; DGTN~\cite{guan2024dynamic}, which models dynamic population relationships; Pop-GCN~\cite{parisot2018disease}, which pioneered population-based graph learning; GATE~\cite{peng2022gate}, which employs graph attention for population analysis; and EV-GCN~\cite{huang2022disease}, which models evolving population structures. Hyperparameters for all competing methods were set according to their original publications.

\section{Results}\label{Result}
\subsection{Classification Performance Comparison}
\subsubsection{10-Fold Cross-Validation}
\begin{table}
	\centering
	\caption{Diagnostic results (mean (std)) for competing methods on ABIDE-I dataset using the AAL116 atlas. The asterisk (*) by the metric indicates significant difference at p < 0.05 level when compared to the suboptimal performance, using either two-sided paired t-test or two-sided Wilcoxon signed-rank test. {\bfseries Bold}: optimal, \underline{Underline}: suboptimal)}\label{result_abide_aal}
	\small
	\setlength{\tabcolsep}{3pt}
	\begin{tabularx}{\textwidth}{>{\centering\arraybackslash}p{1.5cm}>{\centering\arraybackslash}p{1.4cm}>{\centering\arraybackslash}p{3.2cm}>{\centering\arraybackslash}X>{\centering\arraybackslash}X>{\centering\arraybackslash}X}
		\hline
		\textbf{Dataset} & \textbf{Type} & \textbf{Method} & \textbf{ACC(\%)} & \textbf{AUC(\%)} & \textbf{F1(\%)} \\ \hline
		\multirow{15}{*}{ABIDE-I} & \multirow{2}{*}{Traditional} & SVM & 64.22 (0.35) & 64.97 (0.37) & 64.83 (0.45) \\
		& & MLP & 68.89 (0.84) & 72.89 (1.02) & 69.15 (0.95) \\ \cline{2-6}
		& \multirow{5}{*}{B.GCN} & BrainGNN~\cite{li2021braingnn} & 66.76 (3.81) & 69.39 (2.76) & 67.21 (1.94) \\
		& & ContrastPool~\cite{xu2024contrastive} & 70.40 (2.74) & 70.29 (3.48) & 68.03 (2.31) \\
		& & RGTNet~\cite{wang2024residual} & 73.21 (1.86) & 75.10 (2.54) & 72.69 (2.75) \\
		& & DGCN~\cite{zhao2022dynamic} & 71.03 (2.90) & 71.65 (2.98) & 70.44 (2.85) \\
		& & AL-NEGAT~\cite{chen2022adversarial} & 72.37 (3.32) & 74.22 (2.56) & 72.81 (3.15) \\\cline{2-6}
		& \multirow{6}{*}{P.GCN} & InceptionGCN~\cite{kazi2019inceptiongcn} & 69.43 (1.26) & 72.90 (0.97) & 70.25 (1.36) \\
		& & LG-GNN~\cite{zhang2022classification} & 73.27 (1.76) & 75.37 (1.55) & 74.26 (1.94) \\
		& & DGTN~\cite{guan2024dynamic} & 76.71 (1.66) & 79.54 (1.83) & \underline{77.51 (1.72)} \\
		& & Pop-GCN~\cite{parisot2018disease} & 72.63 (0.86) & 73.10 (3.08) & 68.85 (2.15) \\
            & & EV-GCN~\cite{huang2022disease} & \underline{77.14 (0.27)} & \underline{80.73 (0.29)} & 77.18 (0.35)\\
		& & GATE~\cite{peng2022gate} & 73.85 (2.17) & 76.07 (1.64) & 74.92 (2.35) \\
		  \cline{2-6}
		& Ours & I²B-HGNN & \textbf{78.64 (1.58)*} & \textbf{82.03 (2.37)*} & \textbf{80.45 (1.73)*} \\ \hline
	\end{tabularx}
\end{table}

\begin{table}
	\centering
	\caption{Diagnostic results (mean (std)) for competing methods on ABIDE-I dataset using the Schaefer atlas. The asterisk (*) by the metric indicates significant difference at p < 0.05 level when compared to the suboptimal performance, using either two-sided paired t-test or two-sided Wilcoxon signed-rank test. {\bfseries Bold}: optimal, \underline{Underline}: suboptimal)}\label{result_abide_schaefer}
	\small
	\setlength{\tabcolsep}{3pt}
	\begin{tabularx}{\textwidth}{>{\centering\arraybackslash}p{1.5cm}>{\centering\arraybackslash}p{1.4cm}>{\centering\arraybackslash}p{3.2cm}>{\centering\arraybackslash}X>{\centering\arraybackslash}X>{\centering\arraybackslash}X}
		\hline
		\textbf{Dataset} & \textbf{Type} & \textbf{Method} & \textbf{ACC(\%)} & \textbf{AUC(\%)} & \textbf{F1(\%)} \\ \hline
		\multirow{15}{*}{ABIDE-I} & \multirow{2}{*}{Traditional} & SVM & 64.52 (0.72) & 72.13 (0.26) & 66.53 (0.99) \\
		& & MLP & 68.28 (0.42) & 71.99 (1.52) & 69.85 (0.70) \\ \cline{2-6}
		& \multirow{5}{*}{B.GCN} & BrainGNN~\cite{li2021braingnn} & 67.26 (3.41) & 68.09 (3.36) & 68.31 (1.64) \\
		& & ContrastPool~\cite{xu2024contrastive} & 69.60 (3.24) & 71.49 (3.13) & 66.53 (2.76) \\
		& & RGTNet~\cite{wang2024residual} & 74.11 (1.61) & 73.99 (2.94) & 73.29 (2.25) \\
		& & DGCN~\cite{zhao2022dynamic} & 70.33 (3.20) & 73.05 (2.53) & 69.94 (3.40) \\
		& & AL-NEGAT~\cite{chen2022adversarial} & 73.67 (2.92) & 73.62 (2.91) & 73.61 (2.55) \\\cline{2-6}
		& \multirow{6}{*}{P.GCN} & InceptionGCN~\cite{kazi2019inceptiongcn} & 68.43 (1.66) & 73.60 (0.67) & 70.75 (1.61) \\
		& & LG-GNN~\cite{zhang2022classification} & 74.17 (1.41) & 74.17 (2.05) & 73.46 (1.54) \\
		& & DGTN~\cite{guan2024dynamic} & \underline{76.91 (1.96)} & 79.64 (1.58) & \underline{78.21 (2.12)} \\
		& & Pop-GCN~\cite{parisot2018disease} & 74.13 (0.66) & 72.20 (3.68) & 69.25 (1.80) \\
            & & EV-GCN~\cite{huang2022disease} & 76.34 (0.77) & \underline{80.33 (0.14)} & 76.18 (0.80)\\
		& & GATE~\cite{peng2022gate} & 74.45 (1.87) & 74.77 (2.14) & 76.12 (1.95) \\
		  \cline{2-6}
		& Ours & I²B-HGNN & \textbf{79.34 (1.33)*} & \textbf{81.13 (2.77)} & \textbf{79.55 (1.38)*} \\ \hline
	\end{tabularx}
\end{table}

\begin{table}
	\centering
	\caption{Diagnostic results (mean (std)) for competing methods on ADHD-200 dataset using the AAL116 atlas. The asterisk (*) by the metric indicates significant difference at p < 0.05 level when compared to the suboptimal performance, using either two-sided paired t-test or two-sided Wilcoxon signed-rank test. {\bfseries Bold}: optimal, \underline{Underline}: suboptimal)}\label{result_adhd_aal}
	\small
	\setlength{\tabcolsep}{3pt}
	\begin{tabularx}{\textwidth}{>{\centering\arraybackslash}p{1.7cm}>{\centering\arraybackslash}p{1.3cm}>{\centering\arraybackslash}p{3.2cm}>{\centering\arraybackslash}X>{\centering\arraybackslash}X>{\centering\arraybackslash}X}
		\hline
		\textbf{Dataset} & \textbf{Type} & \textbf{Method} & \textbf{ACC(\%)} & \textbf{AUC(\%)} & \textbf{F1(\%)} \\ \hline
		\multirow{15}{*}{ADHD-200} & \multirow{2}{*}{Traditional} & SVM & 65.68 (0.16) & 56.79 (0.21) & 63.45 (0.32) \\
		& & MLP & 67.48 (0.33) & 71.23 (0.40) & 68.65 (0.45) \\ \cline{2-6}
		& \multirow{5}{*}{B.GCN} & BrainGNN~\cite{li2021braingnn} & 65.16 (3.81) & 67.19 (2.86) & 65.71 (2.04) \\
		& & ContrastPool~\cite{xu2024contrastive} & 69.16 (2.85) & 71.19 (2.26) & 67.71 (3.04) \\
		& & RGTNet~\cite{wang2024residual} & 72.19 (1.25) & 75.42 (2.46) & 70.50 (1.49) \\
		& & DGCN~\cite{zhao2022dynamic} & 68.01 (3.69) & 69.25 (4.57) & 67.92 (3.85) \\
		& & AL-NEGAT~\cite{chen2022adversarial} & 68.35 (3.82) & 67.45 (3.69) & 69.42 (3.95) \\
		\cline{2-6}
		& \multirow{6}{*}{P.GCN} & InceptionGCN~\cite{kazi2019inceptiongcn} & 67.76 (2.81) & 70.39 (2.36) & 69.71 (1.94) \\
		& & LG-GNN~\cite{zhang2022classification} & 72.35 (1.48) & 76.12 (1.86) & 74.63 (1.69) \\
		& & DGTN~\cite{guan2024dynamic} & \underline{76.45 (1.98)} & 80.72 (1.96) & 79.63 (2.31) \\
		& & Pop-GCN~\cite{parisot2018disease} & 74.65 (0.32) & 80.92 (1.40) & 72.88 (1.25) \\
            & & EV-GCN~\cite{huang2022disease} & 76.01 (0.96) & \underline{82.33 (1.15)} & \textbf{82.72 (1.08)} \\ 
		& & GATE~\cite{peng2022gate} & 71.40 (0.23) & 76.81 (3.30) & 72.65 (2.85) \\
		\cline{2-6}
		& Ours & I²B-HGNN & \textbf{77.31 (1.14)*} & \textbf{82.63 (1.53)*} & \underline{81.94 (0.98)} \\ \hline
	\end{tabularx}
\end{table}

\begin{table}
	\centering
	\caption{Diagnostic results (mean (std)) for competing methods on ADHD-200 dataset using the Schaefer atlas. The asterisk (*) by the metric indicates significant difference at p < 0.05 level when compared to the suboptimal performance, using either two-sided paired t-test or two-sided Wilcoxon signed-rank test. {\bfseries Bold}: optimal, \underline{Underline}: suboptimal)}\label{result_adhd_schaefer}
	\small
	\setlength{\tabcolsep}{3pt}
	\begin{tabularx}{\textwidth}{>{\centering\arraybackslash}p{1.7cm}>{\centering\arraybackslash}p{1.3cm}>{\centering\arraybackslash}p{3.2cm}>{\centering\arraybackslash}X>{\centering\arraybackslash}X>{\centering\arraybackslash}X}
		\hline
		\textbf{Dataset} & \textbf{Type} & \textbf{Method} & \textbf{ACC(\%)} & \textbf{AUC(\%)} & \textbf{F1(\%)} \\ \hline
		\multirow{15}{*}{ADHD-200} & \multirow{2}{*}{Traditional} & SVM & 64.48 (0.61) & 58.29 (0.11) & 64.25 (0.82) \\
		& & MLP & 68.78 (0.18) & 70.33 (1.00) & 67.15 (0.15) \\ \cline{2-6}
		& \multirow{5}{*}{B.GCN} & BrainGNN~\cite{li2021braingnn} & 65.86 (3.31) & 66.09 (3.21) & 67.51 (2.69) \\
		& & ContrastPool~\cite{xu2024contrastive} & 68.66 (3.40) & 72.79 (1.81) & 68.31 (2.84) \\
		& & RGTNet~\cite{wang2024residual} & 74.09 (1.95) & 74.12 (2.16) & 70.10 (1.89) \\
		& & DGCN~\cite{zhao2022dynamic} & 66.31 (3.09) & 70.05 (4.82) & 69.12 (4.40) \\
		& & AL-NEGAT~\cite{chen2022adversarial} & 69.75 (4.22) & 66.85 (2.99) & 68.42 (3.60) \\
		\cline{2-6}
		& \multirow{6}{*}{P.GCN} & InceptionGCN~\cite{kazi2019inceptiongcn} & 66.26 (3.31) & 71.49 (2.11) & 70.61 (2.69) \\
		& & LG-GNN~\cite{zhang2022classification} & 71.55 (1.13) & 77.82 (2.46) & 75.93 (1.19) \\
		& & DGTN~\cite{guan2024dynamic} & \underline{76.15 (1.58)} & 78.92 (2.26) & 79.93 (1.66) \\
		& & Pop-GCN~\cite{parisot2018disease} & 73.45 (0.77) & \underline{81.32 (1.20)} & 74.48 (1.95) \\
            & & EV-GCN~\cite{huang2022disease} & 76.02 (0.66) & 80.47 (1.70) & \underline{82.00 (0.93)} \\ 
		& & GATE~\cite{peng2022gate} & 73.20 (0.58) & 75.41 (2.80) & 72.15 (3.25) \\
		\cline{2-6}
		& Ours & I²B-HGNN & \textbf{77.10 (1.38)*} & \textbf{82.83 (1.13)*} & \textbf{82.74 (0.51)} \\ \hline
	\end{tabularx}
\end{table}

\begin{figure}
\centering
\includegraphics[width=1\textwidth]{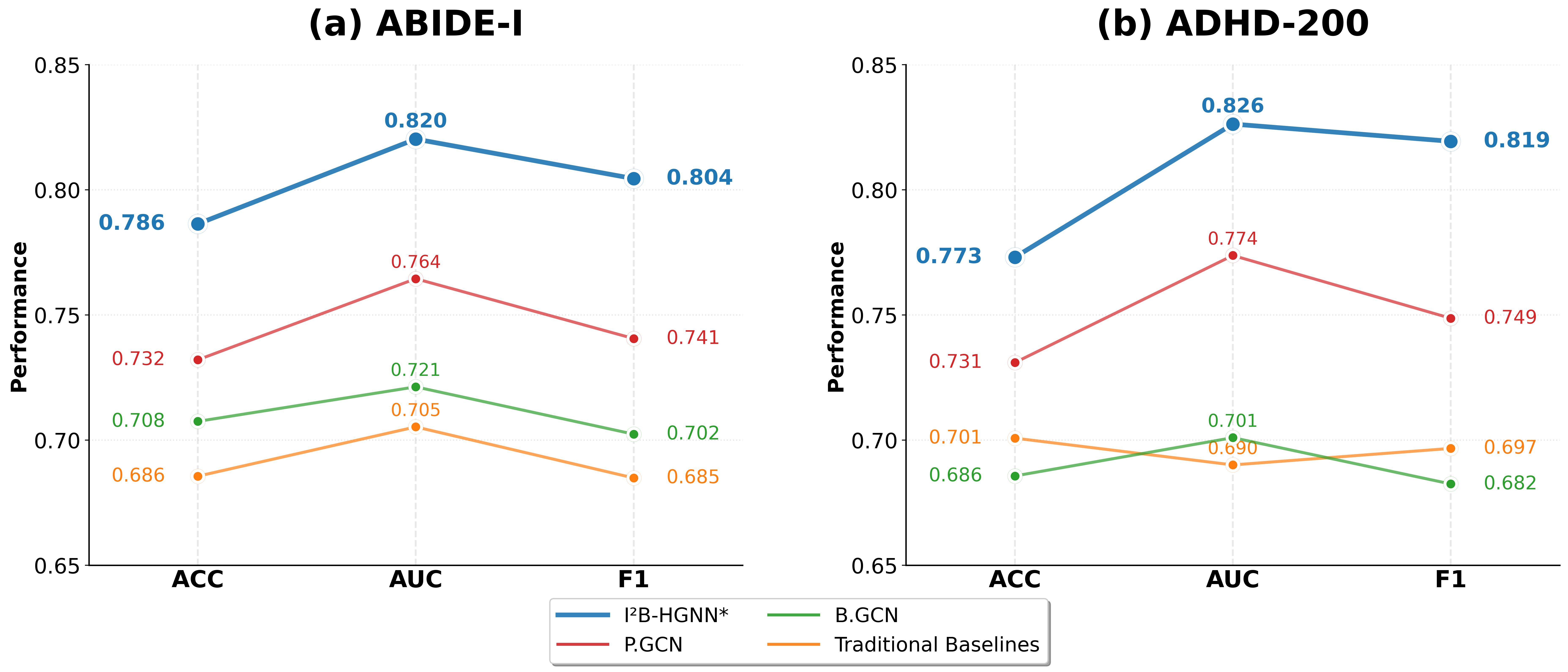}
\caption{Average performance comparison of I²B-HGNN against baseline methods on ABIDE-I and ADHD-200 datasets. The slope graph displays ACC, AUC, and F1 across traditional baselines, B.GCN, P.GCN, and our proposed I²B-HGNN method. I²B-HGNN consistently achieves superior performance across all metrics on both datasets.}
\label{slope_comparison}
\end{figure}
As shown in Tables \ref{result_abide_aal}, \ref{result_abide_schaefer}, \ref{result_adhd_aal} and \ref{result_adhd_schaefer}, I²B-HGNN demonstrates superior performance across all evaluation metrics on both datasets compared to state-of-the-art methods from three categories. On the ABIDE-I dataset, our method achieves 78.64\% accuracy, 82.03\% AUC, and 80.45\% F1 score, outperforming the strongest competing methods EV-GCN by notable margins. On the ADHD-200 dataset, I²B-HGNN attains 77.31\% accuracy, 82.63\% AUC, and 81.94\% F1 score, showing substantial improvements over all baselines. Figure \ref{slope_comparison} further illustrates how I²B-HGNN consistently outperforms the average performance of each method category across all evaluation metrics.

Relative to individual-level brain graph approaches, population-level graph methodologies demonstrate markedly enhanced stability and substantially reduced standard deviations, attributable to their exploitation of population-wide correlation patterns rather than reliance on subject-specific local connectivity characteristics \cite{bessadok2022graph}. Notably, multimodal fusion techniques consistently outperformed their unimodal counterparts, underscoring that neuroimaging data from isolated modalities may lack the requisite discriminative capacity for characterizing complex NDDs \cite{abrol2021deep}. Traditional machine learning approaches exhibited pronounced limitations in this context, reflecting their inherent inability to effectively capture the intricate hierarchical architectures and high-dimensional interdependencies that characterize brain connectivity networks, thereby necessitating the development of specialized graph-based computational frameworks \cite{li2021braingnn}.

\subsubsection{Leave-One-Site-Out Validation}
\begin{table*}
\centering
\caption{Leave-one-site-out cross validation results (ACC) on ABIDE-I dataset using the AAL116 atlas. The asterisk (*) by the metric indicates significant difference at p < 0.05 level when compared to the suboptimal performance, using either two-sided paired t-test or two-sided Wilcoxon signed-rank test. {\bfseries Bold}: optimal, \underline{Underline}: suboptimal.}
\fontsize{7.7}{8.2}\selectfont
\renewcommand\arraystretch{1.05}
\setlength{\tabcolsep}{2.5pt}
\begin{tabularx}{\textwidth}{l*{8}{>{\centering\arraybackslash}X}}
\toprule
\multirow{2}{*}{\fontsize{8.3}{8.3}\selectfont\textbf{Site}} & \multicolumn{3}{c}{\textbf{B.GCN}} & \multicolumn{4}{c}{\textbf{P.GCN}} & \textbf{Ours} \\
\cmidrule(lr){2-4} \cmidrule(lr){5-8} \cmidrule(lr){9-9}
& \textbf{ContrastPool} & \textbf{RGTNet} & \fontsize{7.3}{8.1}\selectfont\textbf{AL-NEGAT} & \textbf{LG-GNN} & \textbf{DGTN} & \textbf{EV-GCN} & \textbf{GATE} & \textbf{I²B-HGNN} \\
\midrule[0.8pt]
CMU & 75.0\% & 75.0\% & 79.2\% & 79.2\% & \underline{83.3\%} & 79.2\% & 79.2\% & \textbf{87.5\%*} \\
CALTECH & 68.4\% & 71.1\% & 68.4\% & 71.1\% & \underline{73.7\%} & 71.1\% & 71.1\% & \textbf{76.3\%} \\
KKI & 69.1\% & \underline{72.7\%} & 70.9\% & 70.9\% & \underline{72.7\%} & \underline{72.7\%} & 70.9\% & \textbf{76.4\%*} \\
LEUVEN & 70.3\% & 73.4\% & \textbf{79.7}\% & 75.0\% & \underline{76.6\%} & \underline{76.6\%} & 73.4\% & \textbf{79.7\%*} \\
MAX\_MUN & 66.7\% & 73.7\% & 71.9\% & 73.7\% & \underline{75.4\%} & \textbf{77.2\%} & 73.7\% & \underline{75.4\%} \\
NYU & 70.7\% & 73.4\% & 71.2\% & 74.5\% & \underline{78.3\%} & 76.6\% & 75.0\% & \textbf{80.4\%} \\
OHSU & 67.9\% & 71.4\% & 71.4\% & 71.4\% & \underline{75.0\%} & \underline{75.0\%} & 71.4\% & \textbf{78.6\%*} \\
OLIN & 72.2\% & 75.0\% & 72.2\% & 75.0\% & \textbf{80.6\%} & \underline{77.8\%} & 75.0\% & \underline{77.8\%} \\
PITT & 71.9\% & 75.4\% & 73.7\% & 75.4\% & 78.9\% & \underline{80.7\%} & 77.2\% & \textbf{82.5\%} \\
SBL & 73.3\% & 76.7\% & \textbf{83.3\%} & \underline{80.0\%} & 76.7\% & 76.7\% & 76.7\% & \textbf{83.3\%*} \\
SDSU & 75.0\% & \underline{77.8\%} & 75.0\% & \underline{77.8\%} & 75.0\% & \underline{77.8\%} & \underline{77.8\%} & \textbf{83.3\%*} \\
STANFORD & 70.0\% & 72.5\% & 70.0\% & 72.5\% & \underline{77.5\%} & \textbf{80.0\%} & 72.5\% & 75.0\% \\
TRINITY & 65.3\% & 69.4\% & 67.3\% & 69.4\% & 71.4\% & \underline{73.5\%} & 69.4\% & \textbf{75.5\%} \\
UCLA & 68.7\% & 72.7\% & 70.7\% & 73.7\% & 76.8\% & \underline{78.8\%} & 74.7\% & \textbf{83.8\%*} \\
UM & \underline{74.5\%} & 68.3\% & 66.9\% & 69.7\% & 72.4\% & \textbf{77.2\%} & 70.3\% & \underline{74.5\%} \\
USM & 71.3\% & 74.3\% & 73.3\% & 75.2\% & 77.2\% & \underline{79.2\%} & 75.2\% & \textbf{85.1\%*} \\
YALE & 75.0\% & 78.6\% & 76.8\% & 78.6\% & \underline{82.1\%} & \textbf{85.7\%} & 78.6\% & \textbf{85.7\%*} \\
\midrule
\textbf{Mean} & 70.9\% & 73.6\% & 73.1\% & 74.3\% & 76.7\% & \underline{77.4\%} & 74.2\% & \textbf{80.0\%} \\
\bottomrule
\end{tabularx}
\label{tab:loso_aal116}
\end{table*}

\begin{table*}
\centering
\caption{Leave-one-site-out cross validation results (ACC) on ABIDE-I dataset using the Schaefer atlas. The asterisk (*) by the metric indicates significant difference at p < 0.05 level when compared to the suboptimal performance, using either two-sided paired t-test or two-sided Wilcoxon signed-rank test. {\bfseries Bold}: optimal, \underline{Underline}: suboptimal.}
\fontsize{7.7}{8.2}\selectfont
\renewcommand\arraystretch{1.05}
\setlength{\tabcolsep}{2.5pt}
\begin{tabularx}{\textwidth}{l*{8}{>{\centering\arraybackslash}X}}
\toprule
\multirow{2}{*}{\fontsize{8.3}{8.3}\selectfont\textbf{Site}} & \multicolumn{3}{c}{\textbf{B.GCN}} & \multicolumn{4}{c}{\textbf{P.GCN}} & \textbf{Ours} \\
\cmidrule(lr){2-4} \cmidrule(lr){5-8} \cmidrule(lr){9-9}
& \textbf{ContrastPool} & \textbf{RGTNet} & \fontsize{7.3}{8.1}\selectfont\textbf{AL-NEGAT} & \textbf{LG-GNN} & \textbf{DGTN} & \textbf{EV-GCN} & \textbf{GATE} & \textbf{I²B-HGNN} \\
\midrule[0.8pt]
CMU & 75.0\% & 79.2\% & 79.2\% & 79.2\% & \underline{83.3\%} & 79.2\% & 79.2\% & \textbf{87.5\%*} \\
CALTECH & 68.4\% & 71.1\% & 68.4\% & 71.1\% & \underline{76.3\%} & 73.7\% & 71.1\% & \textbf{78.9\%*} \\
KKI & 69.1\% & 72.7\% & 70.9\% & 70.9\% & \underline{74.5\%} & \underline{74.5\%} & 70.9\% & \textbf{78.2\%*} \\
LEUVEN & 70.3\% & 73.4\% & \underline{79.7\%} & 75.0\% & 78.1\% & 76.6\% & 73.4\% & \textbf{81.3\%} \\
MAX\_MUN & 66.7\% & 73.7\% & 71.9\% & 73.7\% & \textbf{77.2\%} & \underline{75.4\%} & 73.7\% & \underline{75.4\%} \\
NYU & 70.7\% & 73.4\% & 71.2\% & 74.5\% & \underline{79.3\%} & 78.3\% & 75.0\% & \textbf{81.5\%} \\
OHSU & 67.9\% & 71.4\% & 71.4\% & 71.4\% & \textbf{78.6\%} & \underline{75.0\%} & 71.4\% & \underline{75.0\%} \\
OLIN & 72.2\% & 75.0\% & 72.2\% & 75.0\% & \textbf{80.6\%} & \underline{77.8\%} & 75.0\% & \underline{77.8\%} \\
PITT & 71.6\% & 74.6\% & 73.1\% & 74.6\% & \underline{80.6\%} & 79.1\% & 76.1\% & \textbf{83.6\%*} \\
SBL & 73.3\% & 76.7\% & \textbf{83.3\%} & \underline{80.0\%} & 76.7\% & 76.7\% & 76.7\% & \textbf{83.3\%} \\
SDSU & 75.0\% & \underline{77.8\%} & 75.0\% & \underline{77.8\%} & \underline{77.8\%} & \underline{77.8\%} & \underline{77.8\%} & \textbf{86.1\%*} \\
STANFORD & 70.0\% & 72.5\% & 70.0\% & 72.5\% & \underline{80.0\%} & 77.5\% & 72.5\% & \textbf{82.5\%} \\
TRINITY & 65.3\% & 69.4\% & 67.3\% & 69.4\% & 73.5\% & \underline{75.5\%} & 69.4\% & \textbf{77.6\%} \\
UCLA & 68.7\% & 72.7\% & 70.7\% & 73.7\% & \underline{79.8\%} & 78.8\% & 74.7\% & \textbf{84.8\%*} \\
UM & \underline{74.5\%} & 68.3\% & 66.9\% & 69.7\% & 73.8\% & \textbf{77.2\%} & 70.3\% & 73.8\% \\
USM & 71.3\% & 74.3\% & 73.3\% & 75.2\% & 79.2\% & \underline{80.2\%} & 75.2\% & \textbf{84.2\%*} \\
YALE & 75.0\% & 78.6\% & 76.8\% & 78.6\% & \underline{83.9\%} & 82.1\% & 78.6\% & \textbf{87.5\%*} \\
\midrule
\textbf{Mean} & 70.9\% & 73.8\% & 73.0\% & 74.3\% & \underline{78.6\%} & 77.4\% & 74.2\% & \textbf{81.1\%} \\
\bottomrule
\end{tabularx}
\label{tab:loso_schaefer}
\end{table*}

\begin{table*}[t]
\centering
\caption{Leave-one-site-out cross validation results (ACC) on ADHD-200 dataset using two atlases. The asterisk (*) by the metric indicates significant difference at p < 0.05 level when compared to the suboptimal performance, using either two-sided paired t-test or two-sided Wilcoxon signed-rank test. {\bfseries Bold}: optimal, \underline{Underline}: suboptimal.}
\fontsize{7.7}{8.2}\selectfont
\renewcommand\arraystretch{1.05}
\setlength{\tabcolsep}{2.5pt}
\begin{tabularx}{\textwidth}{l*{8}{>{\centering\arraybackslash}X}}
\toprule
\multirow{2}{*}{\fontsize{8.3}{8.3}\selectfont\textbf{Site}} & \multicolumn{3}{c}{\textbf{B.GCN}} & \multicolumn{4}{c}{\textbf{P.GCN}} & \textbf{Ours} \\
\cmidrule(lr){2-4} \cmidrule(lr){5-8} \cmidrule(lr){9-9}
& \textbf{ContrastPool} & \textbf{RGTNet} & \fontsize{7.3}{8.1}\selectfont\textbf{AL-NEGAT} & \textbf{LG-GNN} & \textbf{DGTN} & \textbf{EV-GCN} & \textbf{GATE} & \textbf{I²B-HGNN} \\
\midrule[0.8pt]
\multicolumn{9}{c}{\textit{\textbf{AAL116 Atlas}}} \\
\midrule[0.4pt]
NYU & 70.1\% & 73.2\% & 69.1\% & 73.7\% & 78.9\% & \underline{79.9\%} & 72.2\% & \textbf{81.4\%} \\
KKI & 68.7\% & 71.1\% & 67.5\% & 71.1\% & \underline{75.9\%} & 74.7\% & 69.9\% & \textbf{78.3\%*} \\
PKU & 71.3\% & 74.5\% & 69.9\% & 74.5\% & \textbf{81.9\%} & 78.7\% & 73.1\% & \underline{81.5\%} \\
UP & 67.4\% & 70.8\% & 66.3\% & \underline{73.0\%} & 72.1\% & 71.9\% & 68.5\% & \textbf{77.5\%*} \\
\cmidrule(lr){1-9}
\textbf{Mean} & 69.4\% & 72.4\% & 68.2\% & 73.1\% & \underline{77.2\%} & 76.3\% & 70.9\% & \textbf{79.7\%} \\
\midrule[0.8pt]
\multicolumn{9}{c}{\textit{\textbf{Schaefer Atlas}}} \\
\midrule[0.4pt]
NYU & 70.1\% & 73.7\% & 69.6\% & 74.2\% & 78.4\% & \underline{78.9\%} & 72.7\% & \textbf{79.9\%} \\
KKI & 68.7\% & 71.1\% & 67.5\% & 71.1\% & \underline{75.9\%} & \underline{75.9\%} & 69.9\% & \textbf{77.1\%*} \\
PKU & 71.3\% & 74.5\% & 70.4\% & 74.5\% & 78.2\% & \underline{78.7\%} & 73.6\% & \textbf{80.1\%} \\
UP & 67.4\% & \underline{72.4\%} & 66.3\% & 71.9\% & 71.9\% & 72.0\% & 68.5\% & \textbf{77.5\%*} \\
\cmidrule(lr){1-9}
\textbf{Mean} & 69.4\% & 72.9\% & 68.5\% & 72.9\% & 76.1\% & \underline{76.4\%} & 71.2\% & \textbf{78.7\%} \\
\bottomrule
\end{tabularx}
\label{tab:loso_adhd}
\end{table*}

To rigorously evaluate the cross-site generalization capability of I²B-HGNN, we conduct leave-one-site-out (LOSO) cross validation on both ABIDE-I and ADHD-200 datasets using AAL116 and Schaefer atlas. LOSO validation trains the model on all sites except one and evaluates it on the completely held-out site, thereby simulating realistic clinical deployment. As presented in Tables~\ref{tab:loso_aal116},~\ref{tab:loso_schaefer}, and~\ref{tab:loso_adhd}, I²B-HGNN consistently achieves the highest mean accuracy across all experimental configurations, substantially outperforming state-of-the-art baseline methods. More importantly, examining individual site performance reveals that I²B-HGNN attains either optimal or suboptimal accuracy in the overwhelming majority of cases across both datasets and parcellation schemes, which demonstrates that our method's advantages are robust and generalizable, independent of atlas choice or site-specific characteristics. For clarity of presentation, we omit methods with mean accuracy below 70\% from these tables, as these methods rarely achieved optimal or suboptimal performance at any individual site.

Several noteworthy patterns emerge from the analysis. First, differential atlas effects across disorders: the finer-grained Schaefer atlas yields superior performance on ABIDE-I, while the coarser AAL116 atlas performs better on ADHD-200. This likely reflects that autism involves subtle, distributed connectivity alterations benefiting from finer resolution, whereas ADHD exhibits more focal dysregulation better captured by anatomical parcellations. Second, performance variation across sites is markedly larger on ABIDE-I compared to ADHD-200, reflecting greater heterogeneity in the ABIDE consortium's scanning protocols and demographic distributions. Sites with reduced accuracy often possess distinctive characteristics such as smaller sample sizes, unique scanner configurations, or atypical demographic profiles that introduce distribution shifts. Third, consistent with the 10-fold cross-validation results, population graph-based approaches generally outperform brain graph-based methods, indicating that leveraging inter-subject relationships significantly enhances cross-site generalization. These comprehensive results establish that our framework maintains high diagnostic accuracy on completely unseen sites, demonstrating its potential for reliable deployment in multi-center clinical settings.

\subsection{Ablation Studies}
\subsubsection{Component-wise Analysis}
\begin{table}[t]
	\centering
	\caption{Ablation studies regarding each key component of our I²B-HGNN.}
	\label{ablation_component}
	\small
	\setlength{\tabcolsep}{3pt}
	\begin{tabularx}{\textwidth}{>{\centering\arraybackslash}p{1.85cm}>{\centering\arraybackslash}p{2.4cm}>{\centering\arraybackslash}p{4.0cm}>{\centering\arraybackslash}X>{\centering\arraybackslash}X>{\centering\arraybackslash}X}
		\hline
		\textbf{Dataset} & \textbf{Component} & \textbf{Configuration} & \textbf{ACC(\%)} & \textbf{AUC(\%)} & \textbf{F1(\%)} \\ \hline
		\multirow{6}{*}{ABIDE-I} & \multirow{2}{*}{IBGraphFormer} & w/o Global Attention & 76.10 & 78.45 & 77.23 \\
		& & w/o $\mathcal{L}_{\text{BIB}}$ & 74.09 & 76.82 & 75.67 \\ \cline{2-6}
		& \multirow{3}{*}{IB-HGAN} & w/o $\mathcal{L}_{\text{HIB}}$ & 73.52 & 75.43 & 74.78 \\
		& & w/o $\mathcal{L}_{\text{struct}}$ & 77.26 & 80.67 & 78.91 \\
		& & w/o $\mathcal{L}_{\text{sparse}}$ & 75.94 & 78.32 & 77.21 \\ \cline{2-6}
		& I²B-HGNN & - & \textbf{78.64} & \textbf{82.03} &\textbf{80.45} \\ \cline{1-6}
		\multirow{6}{*}{ADHD-200} & \multirow{2}{*}{IBGraphFormer} & w/o Global Attention & 75.29 & 79.84 & 78.56 \\
		& & w/o $\mathcal{L}_{\text{BIB}}$ & 74.56 & 79.12 & 77.89 \\ \cline{2-6}
		& \multirow{3}{*}{IB-HGAN} & w/o $\mathcal{L}_{\text{HIB}}$ & 73.84 & 78.67 & 77.34 \\
		& & w/o $\mathcal{L}_{\text{struct}}$ & 76.76 & 81.89 & 80.67 \\
		& & w/o $\mathcal{L}_{\text{sparse}}$ & 75.43 & 80.45 & 79.12 \\ \cline{2-6}
		& I²B-HGNN & - & \textbf{77.31} & \textbf{82.63} & \textbf{81.94} \\ \hline
	\end{tabularx}
\end{table}
To quantify the contribution of each component, we conducted ablation experiments. As shown in Table \ref{ablation_component}, the results demonstrate that removing the heterogeneous population graph loss (w/o $\mathcal{L}_{\text{HIB}}$) causes the most significant performance degradation across all evaluation metrics, highlighting its critical role in enforcing minimal sufficient statistics during multimodal integration, as heuristic fusion strategies cannot adequately capture the complex interdependencies between neuroimaging and demographic data. Removing the BIB-Pooling (w/o $\mathcal{L}_{\text{BIB}}$) results in the second most severe performance decline, validating its effectiveness in biomarker identification and indicating that without this principled compression mechanism, the model fails to distinguish between informative and redundant brain connectivity features essential for effective biomarker identification.

Eliminating the global attention leads to notable performance reduction, confirming that local graph convolutions alone are insufficient for modeling the distributed connectivity patterns spanning multiple functional brain networks crucial. The removal of sparsity regularization (w/o $\mathcal{L}_{\text{sparse}}$) results in moderate but consistent performance drops, where the sparsity loss ensures computational focus on diagnostically relevant demographic relationships. Performance also declined without structural consistency constraints (w/o $\mathcal{L}_{\text{struct}}$), demonstrating their complementary roles in maintaining representation equivalence across isomorphic meta-paths and promoting selective attention to diagnostically relevant pathways. The consistent performance degradation across ablation conditions confirms that our framework achieves optimal diagnostic performance through synergistic integration of information-theoretic principles.

\subsubsection{IB Mechanism Validation}
\begin{figure}[t]
\centering
\includegraphics[width=1\textwidth]{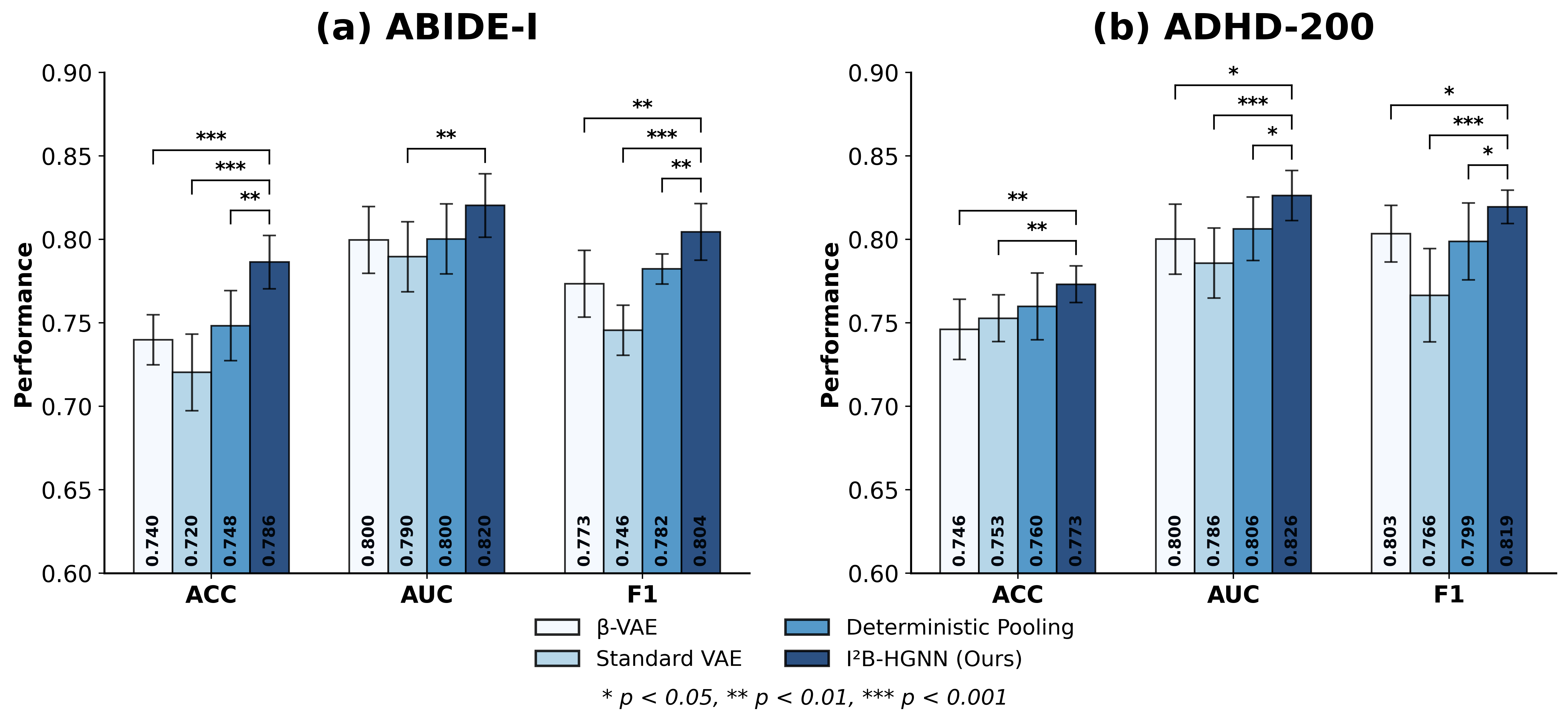}
\caption{Performance comparison of different IB and pooling mechanisms on (a) ABIDE-I and (b) ADHD-200 datasets across multiple evaluation metrics (ACC, AUC, F1). Our IB-guided approach consistently outperforms alternative compression methods.}
\label{fig_ib_comparison}
\end{figure}
We systematically validated the BIB-Pooling by comparing against alternative compression and pooling strategies. Figure~\ref{fig_ib_comparison} provides compelling evidence for the superiority of our information-theoretic approach across all evaluation metrics. We compared against three baselines: $\beta$-VAE \cite{higgins2017beta} which employs a fixed hyperparameter $\beta$ to balance reconstruction quality with representation compactness, Standard VAE \cite{kingma2013auto} which applies unconditional variational compression without task-specific constraints, and Deterministic Pooling \cite{ying2018hierarchical} which uses conventional pooling operations without probabilistic modeling. Our I²B-HGNN consistently outperforms all alternatives across both datasets and evaluation metrics. Standard VAE and $\beta$-VAE suffer from suboptimal compression-performance trade-offs because they treat all features equally during compression and lack adaptive mechanisms to identify and preserve diagnostically relevant biomarkers while potentially discarding task-critical information. Deterministic pooling methods cannot capture the uncertainty inherent in neuroimaging data and fail to provide principled feature selection criteria. Our BIB-Pooling mechanism adaptively preserves features with high mutual information with diagnostic labels while compressing redundant connectivity patterns. The theoretical guarantee of minimal sufficient statistics ensures optimal compression that retains essential diagnostic information.

\subsubsection{Heterogeneous Graph Construction Analysis}
\begin{table}[t]
	\centering
	\caption{Systematic evaluation of heterogeneous graph construction strategies and meta-path configurations.}\label{hetero_analysis}
	\small
	\setlength{\tabcolsep}{3pt}
	\begin{tabularx}{\textwidth}{>{\centering\arraybackslash}p{1.8cm}>{\centering\arraybackslash}p{6.5cm}>{\centering\arraybackslash}X>{\centering\arraybackslash}X>{\centering\arraybackslash}X}
		\hline
		\textbf{Dataset} & \textbf{Method} & \textbf{ACC(\%)} & \textbf{AUC(\%)} & \textbf{F1(\%)} \\ \hline
		\multirow{5}{*}{ABIDE-I} & Homogeneous Population Graph & 73.52 & 75.43 & 74.78 \\
		& Homogeneous (Weighted Combination) & 74.21 & 76.12 & 75.34 \\
		& I²B-HGNN (2 Meta-paths) & 75.67 & 78.32 & 76.89 \\
		& I²B-HGNN (3 Meta-paths) & 77.34 & 81.47 & 78.91 \\
		& I²B-HGNN (4 Meta-paths) & \textbf{78.64} & \textbf{82.03} & \textbf{80.45} \\ \hline
		\multirow{5}{*}{ADHD-200} & Homogeneous Population Graph & 73.84 & 78.67 & 75.23 \\
		& Homogeneous (Weighted Combination) & 74.32 & 79.23 & 75.78 \\
		& I²B-HGNN (2 Meta-paths) & 74.83 & 79.78 & 76.45 \\
		& I²B-HGNN (3 Meta-paths) & 76.12 & 81.45 & 78.67 \\
		& I²B-HGNN (4 Meta-paths) & \textbf{77.31} & \textbf{82.63} & \textbf{81.94} \\ \hline
	\end{tabularx}
\end{table}
To evaluate the effectiveness of our heterogeneous graph construction with structural consistency constraints, we systematically compared I²B-HGNN against homogeneous alternatives and analyzed the contribution of different meta-path configurations. In homogeneous population graph construction, all demographic information is aggregated into a single similarity measure where edge weights between subjects $i$ and $j$ are computed as weighted combinations of individual demographic similarities, treating all demographic relationships uniformly and thus failing to capture the unique characteristics and varying contributions of different demographic factors to diagnostic outcomes. Conversely, our heterogeneous formulation constructs separate meta-path subgraphs $\{\mathcal{G}_k\}_{k=1}^4$ for each demographic attribute, enabling the model to learn distinct attention weights through our IB-guided mechanism and preserve the unique relational patterns inherent in different demographic dimensions \cite{wang2019heterogeneous}.

Table \ref{hetero_analysis} demonstrate substantial advantages of heterogeneous graph modeling over conventional homogeneous approaches, with our method achieving significant performance improvements on both datasets compared to homogeneous population graphs, validating the hypothesis that explicitly modeling different types of demographic relationships through meta-paths captures richer population-level patterns than treating all subjects uniformly \cite{jin2021heterogeneous}. The progressive performance improvement from 2 to 4 meta-paths confirms that each demographic factor contributes unique diagnostic information, with the optimal configuration utilizing all four available demographic attributes to achieve maximum diagnostic performance. The systematic comparison reveals that heterogeneous graph construction provides fundamental advantages by enabling the model to capture complex interplay between different demographic factors while maintaining interpretable and stable learning dynamics through our theoretically grounded attention mechanism and structural consistency constraints.

\subsubsection{Scalability and Computational Efficiency Analysis}
To evaluate the scalability of our proposed I²B-HGNN, we report the hardware consumption of I²B-HGNN under both 10-fold cross-validation and LOSO validation settings. As shown in Table \ref{tab:computational_requirements}, I²B-HGNN demonstrates reasonable computational demands across both evaluation protocols with approximately 1.2M trainable parameters. The higher memory consumption compared to single-modal approaches stems from our dual-level architecture that processes both individual brain networks and population-level heterogeneous graphs. Despite the additional computational overhead, our information bottleneck formulation enables efficient convergence, with early stopping typically triggering at 150-200 epochs. These results demonstrate that I²B-HGNN achieves strong diagnostic performance with practical computational costs suitable for multi-center neuroimaging studies. Future work could explore efficiency improvements through sparse attention mechanisms, meta-path pruning based on mutual information criteria, or knowledge distillation to compressed models for resource-constrained clinical settings.
\begin{table}
\centering
\fontsize{10}{11}\selectfont
\renewcommand{\arraystretch}{1.4}
\caption{Hardware requirements of I²B-HGNN under different validation protocols.}
\begin{tabular}{cccccc}
    \Xhline{1pt}
    \textbf{Dataset} & \textbf{Protocol} & \textbf{Memory (MB)} & \textbf{FLOPs (G)} & \textbf{Time (s)} & \textbf{Time\textsuperscript{*} (s)} \\
    \hline
    \multirow{2}{*}{ABIDE-I} & 10-fold & 8,205 & 0.39 & 91.84 & 82.47 \\
    & LOSO (17 sites) & 9,530 & 0.42 & 98.21 & 77.35 \\
    \hline
    \multirow{2}{*}{ADHD-200} & 10-fold & 7,263 & 0.27 & 51.38 & 39.28 \\
    & LOSO (4 sites) & 7,571 & 0.30 & 54.92 & 42.76 \\
    \Xhline{1pt}
\multicolumn{6}{l}{\footnotesize Note: All metrics represent average values across folds or sites.}\\
\multicolumn{6}{l}{\footnotesize \textbf{Time} denotes training time per epoch; \textbf{Time\textsuperscript{*}} denotes training time with early stopping.}\\
\end{tabular}
\label{tab:computational_requirements}
\end{table}

\section{Discussion}\label{Discussion}
\subsection{Visualization of I²B-HGNN's Feature Representation}
\begin{figure}[t]
  \centering
  \includegraphics[width=\textwidth]{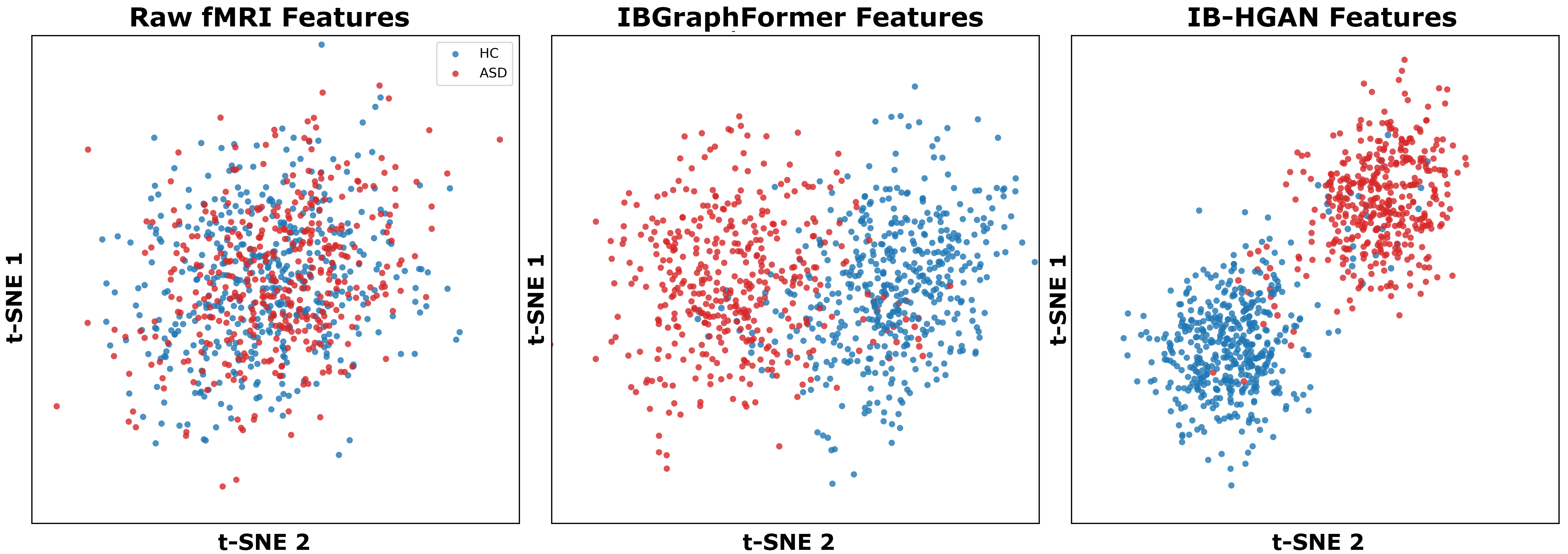}
  \caption{t‑SNE visualization of feature representations on the ABIDE-I dataset.}
  \label{tsne}
\end{figure}
To evaluate I²B-HGNN's capability in learning discriminative representations, we utilized t-SNE \cite{maaten2008visualizing} visualization to project high-dimensional features into two-dimensional space for the ABIDE-I dataset. Figure \ref{tsne} demonstrates the enhancement of feature discriminability across our framework's components. The raw fMRI features exhibit substantial overlap between HC and ASD groups, indicating limited discriminative capacity of original connectivity patterns. Following IBGraphFormer processing, the learned biomarker representations show improved clustering with reduced inter-class overlap, validating the effectiveness of our distribution-aware global attention mechanism combined with BIB-Pooling in extracting diagnostically relevant brain connectivity patterns. The final IB-HGAN features achieve optimal separation, forming two distinct, well-separated clusters with minimal boundary ambiguity. The clear cluster separation with low intra-class and high inter-class dispersion indicates that the multimodal features learned by I²B-HGNN exhibit significant discriminative power.

\subsection{IB Regularization Parameter Analysis}
\begin{figure}
\centering
	\includegraphics[width=1\textwidth]{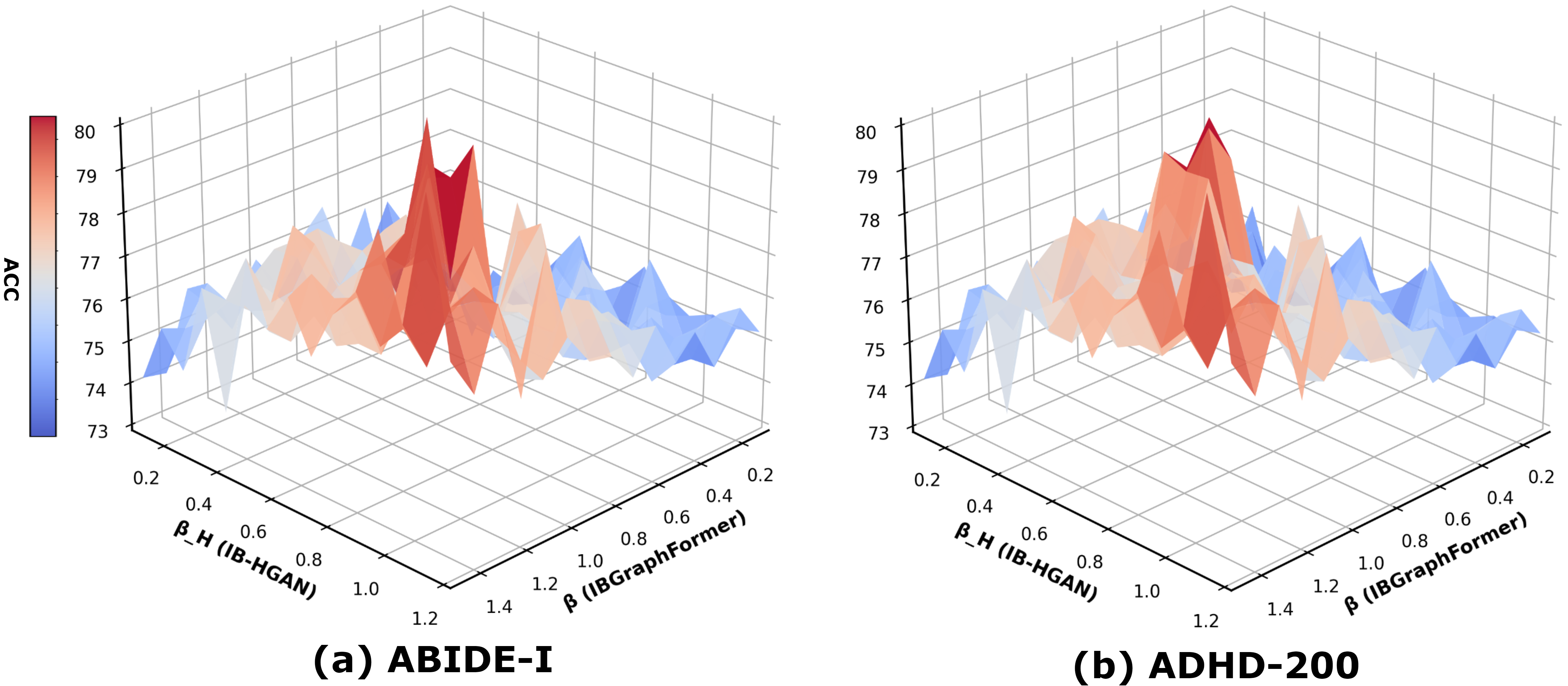}
	\caption{The 3D heatmap of the classification results on the ABIDE-I and ADHD-200 datasets with different IB regularization parameters $\beta$ for IBGraphFormer and $\beta_H$ for IB-HGAN.} \label{IB}
\end{figure}
We conduct systematic experiments by adjusting the compression strength parameters $\beta$ for IBGraphFormer and $\beta_H$ for IB-HGAN to identify optimal combinations for information compression parameters. Displayed in Figure \ref{IB}, clear optimal regions emerge with pronounced sensitivity to parameter combinations. The ABIDE-I dataset demonstrates peak performance at $\beta = 0.8$ and $\beta_H = 0.5$, while ADHD-200 achieves optimal results at $\beta = 1.0$ and $\beta_H = 0.8$, indicating dataset-specific optimal compression-prediction trade-offs. Low $\beta$ values result in insufficient compression, allowing redundant connectivity patterns to dominate learned representations and compromising generalization capacity. Conversely, excessive compression eliminates diagnostically relevant neural patterns, particularly distributed connectivity signatures essential for NDDs characterization. The asymmetric performance landscapes suggest IBGraphFormer requires stronger regularization than IB-HGAN, reflecting the higher dimensionality and complexity of individual brain connectivity patterns compared to population-level demographic relationships \cite{sun2022structural}. While optimal parameters exhibit dataset-specific variations, we observe generalizable selection principles that facilitate systematic parameter tuning. We recommend starting with moderate values ($\beta = 0.8$, $\beta_H = 0.5$) as initial candidates, then prioritizing combinations that yield stable performance across multiple folds rather than peak single-fold performance, while verifying that learned biomarkers align with established neurobiological knowledge. For new datasets with similar characteristics, these default values provide strong starting points, with fine-tuning typically converging within a narrow range ($\pm 0.2$ for both parameters).

\subsection{Interpretability Analysis}
\subsubsection{Biomarker Visualization}
\begin{figure}
\centering
	\includegraphics[width=0.82\textwidth]{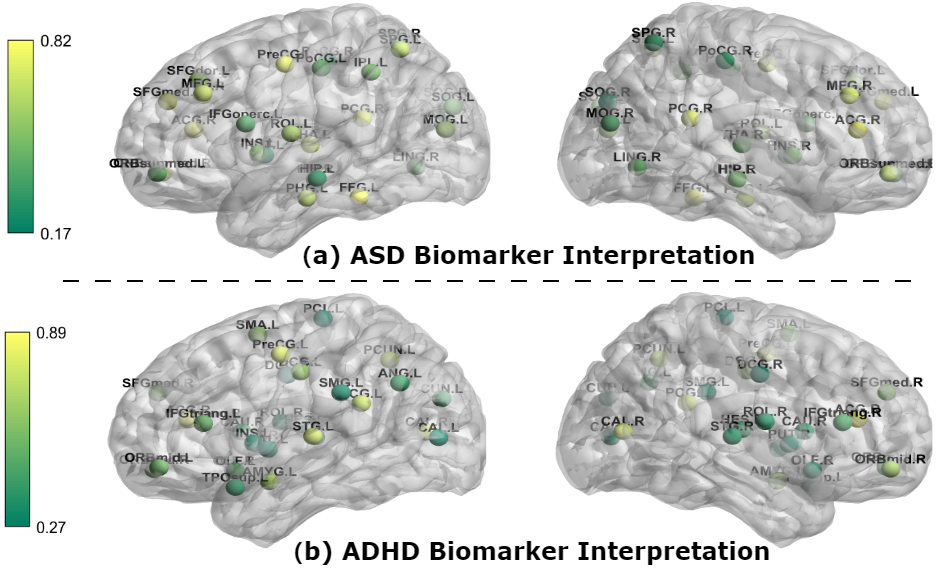}
	\caption{The most important brain regions for NDDs classification.} \label{biomarker}
\end{figure}
Figure \ref{biomarker} visualize the top 30 informative ROIs identified by the BIB-Pooling layer for ADHD and ASD, using IB principle-based normalized mutual information quantification. For ASD classification, biomarkers with high scores were primarily located in the fusiform gyrus (FFG.L), anterior cingulate cortex (ACG.R), precentral gyrus (PreCG.L), superior frontal gyrus medial (SFGmed.L), posterior cingulate gyrus (PCG.R), thalamus (THA.L), and bilateral middle frontal gyrus regions \cite{thapar2017neurodevelopmental,berto2022association}. These findings align with previous neuroimaging studies, where the fusiform gyrus has been consistently reported as a critical region for face processing deficits in ASD \cite{kong2019classification}. The anterior cingulate cortex and frontal regions are associated with social cognition, executive control, and emotional regulation, while the thalamus serves as a relay station for sensory and motor information processing \cite{berto2022association}. All these regions are fundamentally involved in social communication, sensory processing, and executive functions, domains where individuals with ASD exhibit characteristic impairments \cite{thapar2017neurodevelopmental}.

For ADHD classification, the highest-scoring biomarkers were identified in the anterior cingulate cortex (ACG.R), precentral gyrus (PreCG.L), calcarine cortex (CAL.R), superior temporal gyrus (STG.L), posterior cingulate gyrus (PCG.L), precuneus (PCUN.L), and amygdala (AMYG.L) regions \cite{cortese2012toward,wang2023multiple}. Previous studies have demonstrated that individuals with ADHD exhibit functional abnormalities in fronto-striatal circuits, particularly involving the anterior cingulate cortex and prefrontal regions that are crucial for attention control and inhibitory processing. The precentral gyrus, being part of the primary motor cortex, relates to the motor hyperactivity symptoms characteristic of ADHD \cite{thapar2017neurodevelopmental}. The amygdala's involvement reflects emotional dysregulation commonly observed in ADHD patients, while the precuneus and posterior cingulate cortex are key components of the default mode network, which shows altered activation patterns in attention deficit disorders. The superior temporal gyrus contributes to auditory processing and language comprehension, functions often impaired in ADHD populations \cite{bush2010attention}. All identified regions demonstrate strong correlations with the core symptom dimensions of ADHD, including inattention, hyperactivity, and impulsivity.

\begin{figure}
\centering
\includegraphics[width=0.95\textwidth]{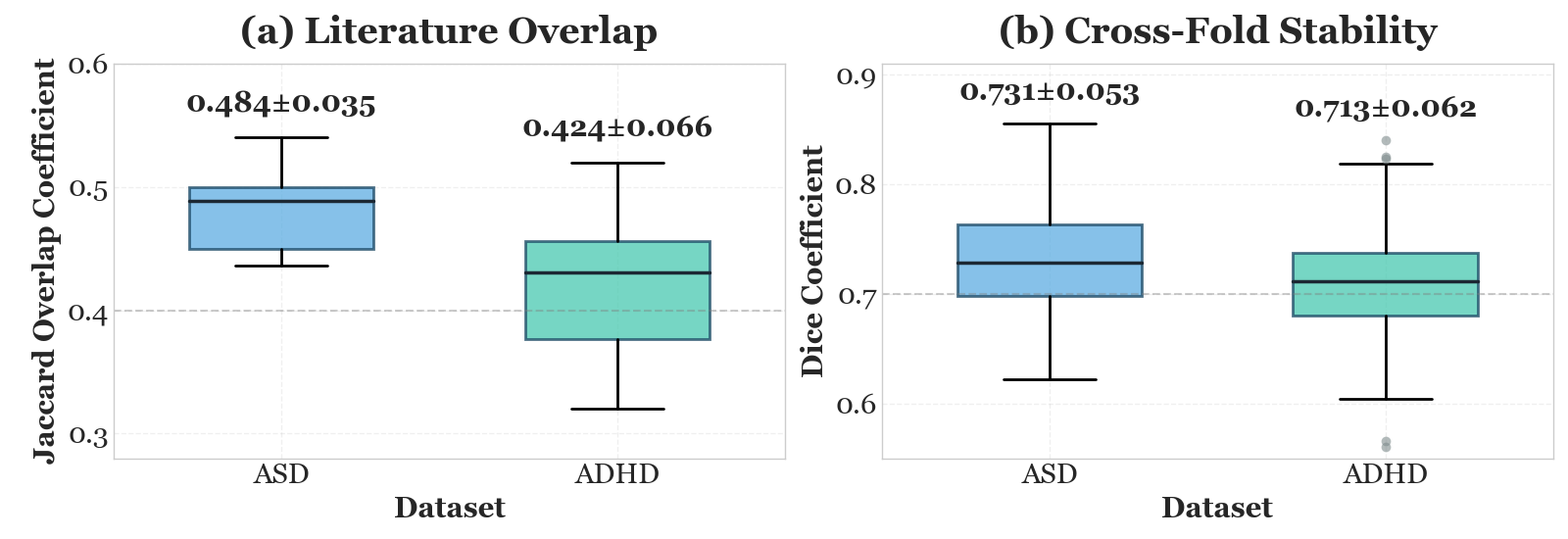}
\caption{Quantitative validation of biomarker reliability. (a) Jaccard overlap with literature-reported regions. (b) Cross-fold stability measured by Dice coefficients. Consistently high metrics confirm reliable biomarker identification for both disorders.}
\label{fig:quantitative_validation}
\end{figure}

To quantitatively assess the clinical relevance and robustness of the identified biomarkers, we conducted two complementary validation analyses using the AAL116 brain atlas parcellation. We first evaluated literature concordance by computing Jaccard overlap coefficients between the top-30 ROIs identified by BIB-Pooling and established biomarkers compiled from large-scale neuroimaging studies \cite{di2014autism,cortese2012toward,berto2022association,wang2023multiple,di2009functional}, comprising 45 reference regions for ASD and 42 for ADHD in Table~\ref{tab:reference_regions}. Additionally, we assessed cross-fold stability by computing Dice coefficients between biomarker sets identified across different folds of the 10-fold cross-validation. As illustrated in Figure~\ref{fig:quantitative_validation}, our method demonstrates both strong literature concordance and high cross-fold consistency across both disorders. For ASD, approximately 70\% of the identified ROIs overlap with literature-reported biomarkers (Jaccard: 0.484±0.035), while for ADHD, the overlap remains substantial despite greater heterogeneity in reported regions (Jaccard: 0.424±0.066). The cross-fold stability analysis further reveals that core brain regions are consistently identified across different data partitions in both cohorts (Dice: 0.731±0.053 for ASD, 0.713±0.062 for ADHD), demonstrating comparable robustness between datasets.

\begin{table*}[t]
\centering
\caption{Literature-derived reference biomarker regions from large-scale neuroimaging studies \cite{di2014autism,cortese2012toward,berto2022association,wang2023multiple,di2009functional}, mapped to AAL116 atlas.}
\fontsize{8}{9}\selectfont
\renewcommand\arraystretch{1.05}
\setlength{\tabcolsep}{3.5pt}
\begin{tabularx}{\textwidth}{>{\raggedright\arraybackslash}X|>{\raggedright\arraybackslash}X}
\toprule
\textbf{ASD Reference Regions (n=45)} & \textbf{ADHD Reference Regions (n=42)} \\
\midrule[0.8pt]
FFG.L, ACG.R, PreCG.L, SFGmed.L, & ACG.R, PreCG.L, CAL.R, STG.L, \\
PCG.R, THA.L, MFG.L, MFG.R, & PCG.L, PCUN.L, AMYG.L, ORBmid.R, \\
ORBsupmed.R, PHG.L, SPG.L, ROL.L, & DCG.L, SMA.L, SFGmed.R, ORBmid.L, \\
IFGoperc.L, SFGdor.L, PoCG.L, INS.L, & IFGtriang.L, OLF.L, IFGtriang.R, CUN.L, \\
MOG.L, SOG.R, INS.R, IPL.L, & ANG.L, HES.R, TPOsup.L, CAU.R, \\
SOG.L, HIP.R, FFG.R, ACG.L, & PUT.R, PAL.L, CAL.L, DCG.R, \\
PreCG.R, SFGmed.R, PCG.L, THA.R, & PCL.L, ACG.L, PreCG.R, PCUN.R, \\
PHG.R, SPG.R, ROL.R, IFGoperc.R, & AMYG.R, SFGmed.L, HES.L, TPOsup.R, \\
SFGdor.R, PoCG.R, LING.L, IPL.R, & CAU.L, SMG.R, INS.R, PUT.L, \\
HIP.L, PUT.L, STG.L, STG.R, & PAL.R, THA.L, THA.R, PCG.R, \\
MTG.L, AMYG.L, CAU.L, MOG.R, PUT.R & ANG.R, SPG.L \\
\bottomrule
\end{tabularx}
\label{tab:reference_regions}
\end{table*}

\subsubsection{Cross-modal Information Analysis}
\begin{figure}[t]
\centering
	\includegraphics[width=1\textwidth]{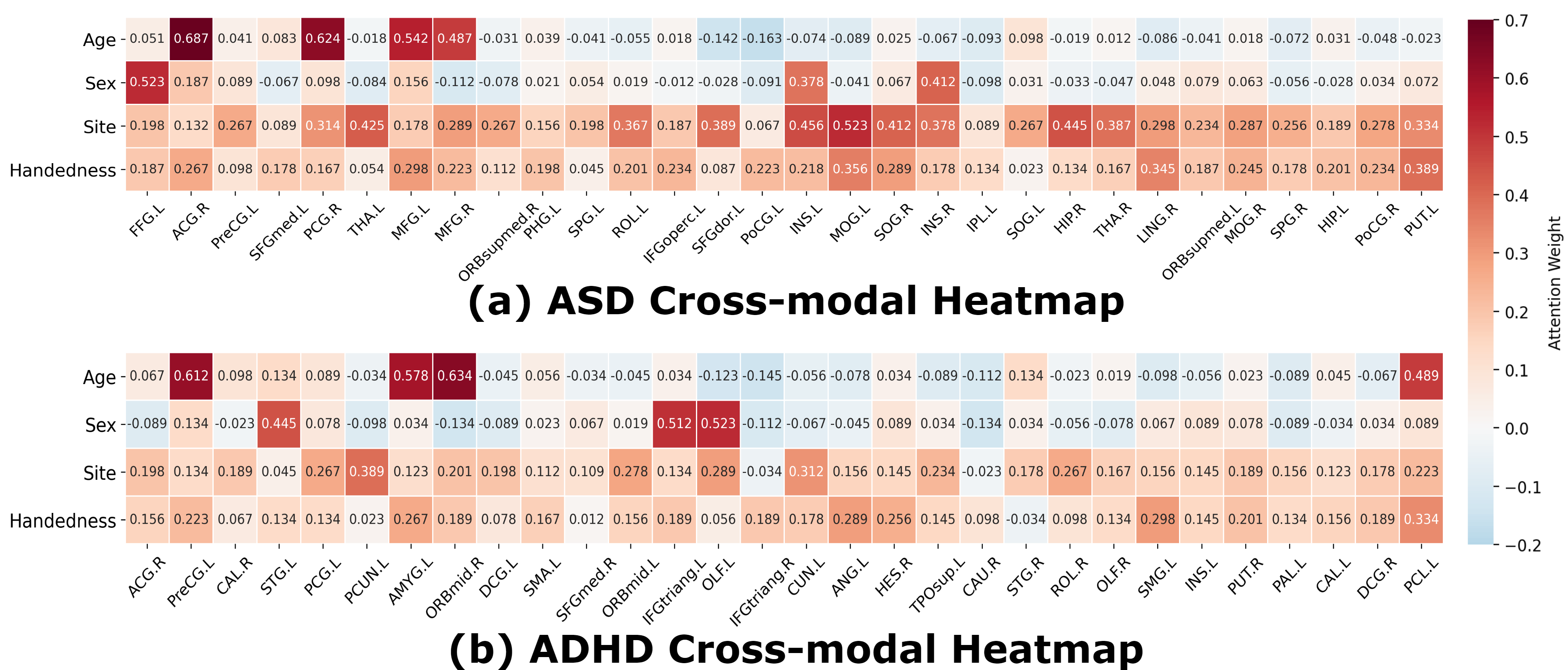}
	\caption{Cross-modal information analysis revealing interaction patterns between demographic factors and brain regions.} \label{explain}
\end{figure}

\begin{figure}[t]
\centering
	\includegraphics[width=0.85\textwidth]{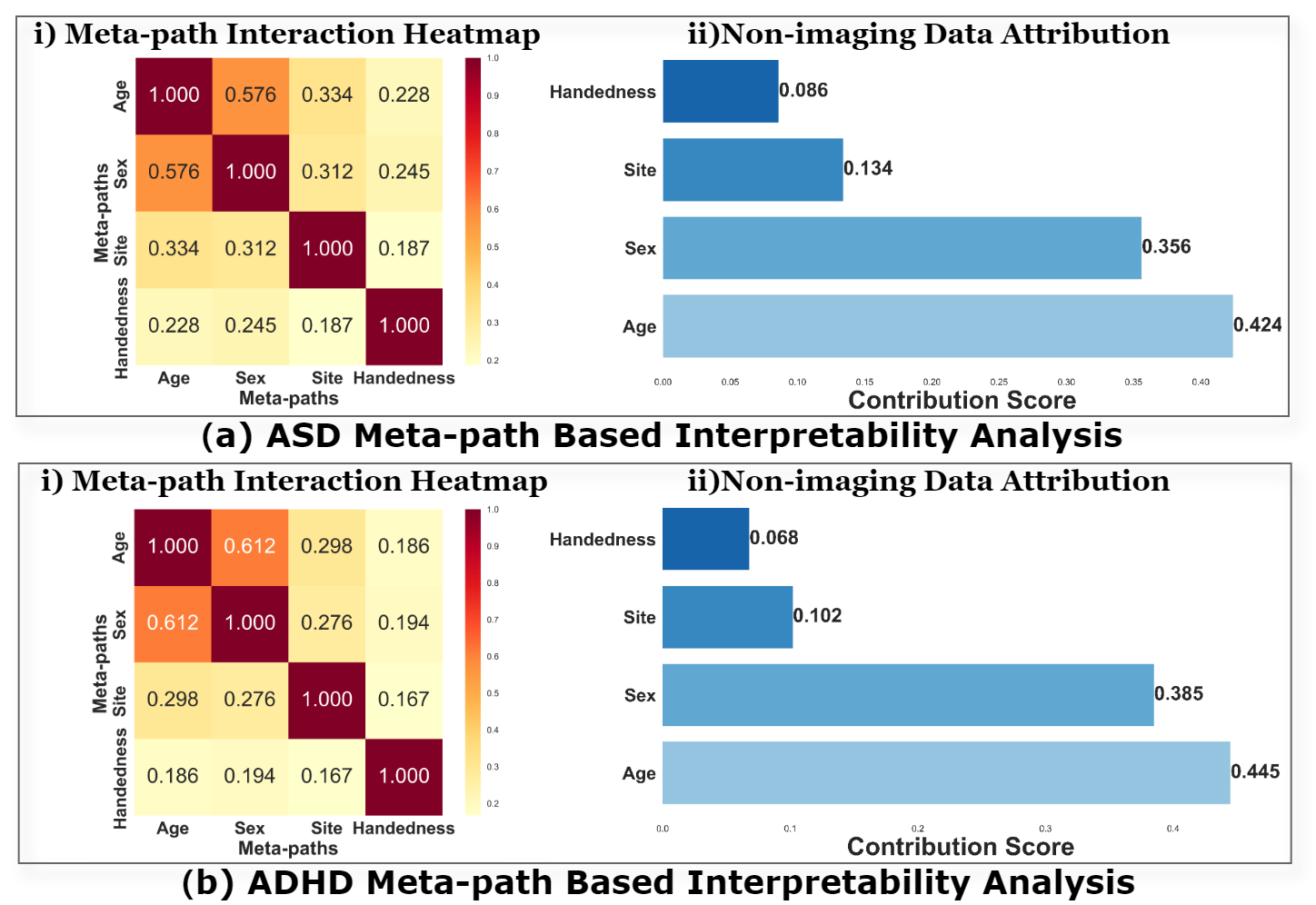}
	\caption{Meta-path interaction analysis and demographic attribution for NDDs diagnosis. (i) Meta-path interaction heatmaps showing correlations between demographic meta-paths for (a) ASD and (b) ADHD datasets. (ii) Non-imaging data attribution showing contribution scores of each demographic factor to diagnostic decisions.} \label{metapath}
\end{figure}
Figure \ref{explain} reveals distinct neurobiological signatures in ASD and ADHD through interactive patterns of cross-modal information flow. The interpretability results demonstrate how I²B-HGNN achieves theoretically-principled explanations by visualizing multimodal feature interactions across demographic, clinical, and neuroimaging modalities. Figure \ref{explain} (a) shows that ASD exhibits stronger site-related correlations with posterior brain regions, particularly evident in the fusiform gyrus (FFG.L) which displays pronounced sex-related attention weights. This pattern aligns with findings that fusiform gyrus alterations provide a neurobiological basis for face processing deficits characteristic of autism \cite{berto2022association,kong2019classification}. The enhanced multimodal interactions across occipital and temporal regions (MOG.L, SOG.R, SPG.L) reflect the posterior brain network alterations where demographic factors modulate neural connectivity through cross-modal information channels \cite{thapar2017neurodevelopmental}. Figure \ref{explain} (b) demonstrates that ADHD exhibits pronounced age-sex interactions with frontal-insular networks, confirming that the anterior cingulate cortex shows consistent alterations during attention tasks across large-scale studies \cite{cortese2012toward,norman2024subcortico}. The precentral gyrus (PreCG.L) shows strong age-related attention weights, supporting evidence that cortical differences reflect developmental trajectories in ADHD \cite{wang2023multiple}. Notably, ADHD displays more concentrated frontal-executive network interactions compared to ASD's distributed posterior-temporal patterns, reflecting the distinct pathophysiological mechanisms underlying these disorders \cite{kofler2024executive}. Meta-path interactions demonstrate how demographic factors influence diagnosis through graph isomorphism-constrained information channels, with ADHD showing pronounced age-sex interactions with frontal-insular networks \cite{sigar2025age,santos2022male}, while ASD displays more posterior-focused multimodal coupling patterns. This interpretability framework preserves crucial cross-modal relationships underlying the pathophysiology of NDDs \cite{bush2010attention}.

\subsubsection{Meta-path Interaction and Attribution Analysis}
Figure \ref{metapath} demonstrates the effectiveness of our IB-guided meta-path attention mechanism in capturing complex demographic relationships and their diagnostic contributions. The meta-path interaction analysis reveals that age-sex correlations exhibit stronger interdependency in ADHD compared to ASD, indicating more pronounced developmental sex-dependent trajectories in ADHD pathophysiology, while ASD demonstrates more balanced demographic interactions with reduced inter-factor dependencies \cite{sigar2025age,santos2022male}. The attribution analysis shows that both age and sex serve as primary diagnostic indicators, with ADHD exhibiting slightly enhanced age dependency reflecting the developmental nature of attention-related symptoms, whereas both disorders demonstrate substantial sex-related contributions consistent with the documented male predominance in neurodevelopmental conditions \cite{thapar2017neurodevelopmental}. Site effects maintain moderate differential influence across disorders, successfully controlled by our heterogeneous framework, while handedness demonstrates relatively lower but non-negligible contributions to diagnostic decisions \cite{kofler2024executive}. These quantitative demographic interaction patterns align with clinical observations that ADHD symptoms often exhibit stronger developmental trajectories compared to the more persistent characteristics of ASD, supporting our model's capacity to capture neurobiologically meaningful population-level diagnostic signatures through theoretically-principled multimodal integration \cite{sadozai2024executive}.

\subsubsection{Clinical Interpretability and Potential Applications}
The interpretability features demonstrated in preceding analyses translate directly to clinical utility. The identified biomarkers (Figure \ref{biomarker}) align with established neurocircuitry models, where fusiform gyrus alterations in ASD correspond to face processing deficits and anterior cingulate/frontal network alterations in ADHD relate to attention control impairments, allowing clinicians to connect imaging findings with behavioral observations. The cross-modal information analysis (Figure \ref{explain}) reveals how demographic factors modulate neural signatures, enabling age-appropriate interpretation of connectivity patterns, while meta-path attribution (Figure \ref{metapath}) provides transparent consideration of non-imaging factors crucial for building clinical trust. These interpretable outputs enable practical applications including early screening to identify high-risk children warranting comprehensive evaluation, differential diagnosis to distinguish between ASD, ADHD, and comorbid presentations in ambiguous cases, treatment planning through personalized biomarker profiles that inform targeted interventions, and longitudinal monitoring to track treatment response. The population graph approach further facilitates multi-center collaborative research while accounting for site-specific variability. While these applications demonstrate significant clinical potential, real-world implementation requires addressing the generalizability, validation, and adoption challenges discussed in the following section.

\subsection{Limitation and future work}
\subsubsection{Generalizability to Clinical Settings}
While our multi-site validation demonstrates technical robustness across diverse research scanning protocols, translation to real-world clinical deployment requires additional considerations. Our framework was developed on research-grade datasets with controlled acquisition parameters. Clinical implementation requires prospective validation on clinically-acquired data with greater heterogeneity, assessment on scanners and protocols not represented in training data, and evaluation in diverse clinical populations differing in symptom severity, comorbidities, and demographic distributions. Real-world clinical settings present challenges of variable scan quality from patient motion, diverse scanner manufacturers and field strengths, and integration with existing clinical workflows and electronic health records. Longitudinal studies assessing model stability across software updates and hardware variations are needed to ensure long-term clinical reliability.

\subsubsection{Clinical Application and Future Directions}
External validation on independent prospectively collected clinical cohorts remains essential for establishing clinical utility. Future work should include validation across broader demographic populations, diverse age ranges, and international cohorts with varying diagnostic criteria and cultural contexts. Enhancing clinical interpretability requires developing case-based reasoning systems connecting identified biomarkers to specific symptom profiles and treatment responses. A critical future direction is developing clinical diagnostic assistance software with user-friendly interfaces providing predictions with confidence estimates, ROI visualizations with network mappings, consultation histories, and similar case retrieval for differential diagnosis. This requires patient data security compliance, quality control modules for scan quality assessment, calibrated uncertainty estimates, and clinician feedback mechanisms for model refinement through clinical partnerships.

\section{Conclusion}\label{conclusion}
In this paper, we present I²B-HGNN, a novel IB-guided framework for interpretable analysis of neurodevelopmental disorders (NDDs), including ASD and ADHD. Our approach innovatively leverages the IB principle to guide both local functional connectivity pattern learning and global multimodal integration, while addressing the critical accuracy-interpretability trade-off through a theoretically grounded progressive architecture. The framework demonstrates how IB principles can effectively guide heterogeneous graph learning for interpretable NDDs diagnosis, enabling simultaneous biomarker identification and non-imaging feature attribution. Experimental
results confirm that I²B-HGNN achieves both high diagnostic accuracy and comprehensive model interpretability. This establishes a new paradigm for principled heterogeneous graph learning in clinical neuroscience applications.

\section*{Acknowledgments}
This work was supported by The Hong Kong Polytechnic University Start-up Fund (Project ID: P0053210), The Hong Kong Polytechnic University Faculty Reserve Fund (Project ID: P0053738), an internal grant from The Hong Kong Polytechnic University (Project ID: P0048377), The Hong Kong Polytechnic University Departmental Collaborative Research Fund (Project ID: P0056428), The Hong Kong Polytechnic University Collaborative Research with World-leading Research Groups Fund (Project ID: P0058097) and Research Grants Council Collaborative Research Fund (Project ID: 5033-24G) and in part by Shenzhen-Hong Kong Institute of Brain Science-Shenzhen Fundamental Research Institutions (Project ID: 2023SHIBS0003).

\bibliographystyle{unsrt}  
\bibliography{ref}
\end{document}